\documentclass[9pt,twocolumn,twoside]{osajnl_patched}
\journal{ol} 

\usepackage{times}
\usepackage{dblfloatfix}
\usepackage{amsmath}
\usepackage{amssymb}
\usepackage{multirow}
\usepackage{multirow}
\usepackage{mathtools}
\usepackage{float}
\usepackage{xcolor}
\usepackage{url}
\usepackage{bbm}
\usepackage{algorithm}
\usepackage{algpseudocode}
\usepackage{pbox}
\usepackage{placeins}
\usepackage{bm}
\usepackage{acronym}
\usepackage{enumitem}

\usepackage{booktabs}
\usepackage{graphicx}
\usepackage{caption}
\usepackage{subcaption}
\usepackage{newfloat}
\usepackage{times}
\usepackage{amsmath}
\usepackage{amssymb}
\usepackage{caption}
\usepackage{bm}
\usepackage{url}
\usepackage{hyperref}
\usepackage[capitalise]{cleveref}
\usepackage{amsmath}
\usepackage{empheq}
\usepackage{multirow}
\usepackage{bm}
\usepackage{graphicx}
\usepackage{amssymb}
\usepackage{pifont} 
\usepackage{booktabs,siunitx,makecell}
\setboolean{shortarticle}{false}

\newcommand{\xmark}{\ding{55}}
\usepackage{newfloat}
\usepackage{multicol}

\newcommand{\q}{\mathbf{q}}

\newcommand{\J}{\mathbf{J}}

\newcommand{\M}{\mathbf{M}}

\newcommand{\F}{\mathbf{F}}
\newcommand{\I}{\mathbf{I}}
\newcommand{\h}{\mathbf{h}}
\renewcommand{\v}{\mathbf{v}}
\renewcommand{\S}{\mathbf{S}}

\newcommand{\G}{\mathbf{G}}
\newcommand\mat[1]{\begin{bmatrix}#1\end{bmatrix}} 

\title{\bfseries \boldmath ZEST: Zero-shot Embodied Skill Transfer for \\Athletic Robot Control} 

\author[1$^\dagger$]{Jean-Pierre Sleiman}
\author[1$^\dagger$]{He Li}
\author[2$^\dagger$]{Alphonsus Adu-Bredu}
\author[2$^\dagger$]{Robin Deits}
\author[2$^\dagger$]{Arun Kumar}

\author[2$^\ddagger$]{Kevin Bergamin}
\author[2$^\ddagger$]{Mohak Bhardwaj}
\author[1$^\ddagger$]{Scott Biddlestone}
\author[1$^\ddagger$]{Nicola Burger}
\author[1$^\ddagger$]{Matthew A. Estrada}
\author[1$^\ddagger$]{Francesco Iacobelli}
\author[2$^\ddagger$]{Twan Koolen}
\author[2$^\ddagger$]{Alexander Lambert}
\author[1$^\ddagger$]{Erica Lin}
\author[1$^\ddagger$]{M. Eva Mungai}
\author[1$^\ddagger$]{Zach Nobles}
\author[2$^\ddagger$]{Shane Rozen-Levy}
\author[1$^\ddagger$]{Yuyao Shi}
\author[1$^\ddagger$]{Jiashun Wang}
\author[2$^\ddagger$]{Jakob Welner}
\author[1$^\ddagger$]{Fangzhou Yu}
\author[1$^\ddagger$]{Mike Zhang}

\author[1$^{\ddagger\ddagger}$]{Alfred Rizzi}
\author[1$^{\ddagger\ddagger}$]{Jessica Hodgins}
\author[1$^{\ddagger\ddagger}$]{Sylvain Bertrand}
\author[2$^{\ddagger\ddagger}$]{Yeuhi Abe}
\author[2$^{\ddagger\ddagger}$]{Scott Kuindersma}
\author[1$^{\ddagger\ddagger}\ast$]{Farbod Farshidian}

\affil[1]{RAI Institute, USA}
\affil[2]{Boston Dynamics, USA}

\affil[$^\dagger$]{Core contributors are listed in order of relative contribution.}
\affil[$^\ddagger$]{Additional contributors are listed alphabetically.}
\affil[$^{\ddagger\ddagger}$]{Project leads are listed in reverse order of relative contribution.}
\affil[$^\ast$]{Corresponding author. Email: \texttt{ffarshidian@rai-inst.com}}

\begin{abstract}

\textbf{{Achieving robust, human-like whole-body control on humanoid robots for agile, contact-rich behaviors remains a central challenge, demanding heavy per-skill engineering and a brittle process of tuning controllers. 
We introduce \emph{ZEST} (Zero-shot Embodied Skill Transfer), a streamlined motion-imitation framework that trains policies via reinforcement learning from diverse sources---high-fidelity motion capture, noisy monocular video, and non-physics-constrained animation---and deploys them to hardware zero-shot. 
\emph{ZEST} generalizes across behaviors and platforms while avoiding contact labels, reference or observation windows, state estimators, and extensive reward shaping.
Its training pipeline combines adaptive sampling, which focuses training on difficult motion segments, and an automatic curriculum using a model-based assistive wrench, together enabling dynamic, long-horizon maneuvers.
We further provide a procedure for selecting joint-level gains from approximate analytical armature values for closed-chain actuators, along with a refined model of actuators.
Trained entirely in simulation with moderate domain randomization, \emph{ZEST} demonstrates remarkable generality. 
On Boston Dynamics' Atlas humanoid, \emph{ZEST} learns dynamic, multi-contact skills (e.g., army crawl, breakdancing) from motion capture. It transfers expressive dance and scene-interaction skills, such as box-climbing, directly from videos to Atlas and the Unitree G1\footnote{The work on Boston Dynamics' Atlas and Spot involved all authors, while authors affiliated with the RAI Institute completed the work on the Unitree G1 robot.}. Furthermore, it extends across morphologies to the Spot quadruped, enabling acrobatics, such as a continuous backflip, through animation. 
Together, these results demonstrate robust zero-shot deployment across heterogeneous data sources and embodiments, establishing \emph{ZEST} as a scalable interface between biological movements and their robotic counterparts.
}}

\end{abstract}

\DeclareFloatingEnvironment{Movie}

\begin{document}

\maketitle


\section*{INTRODUCTION}

\noindent The design of humanoid robots is motivated by a simple fact: our environments are built around the shape, scale, and motion capabilities of the human body. A robot with a similar form can, in principle, operate within these environments without requiring extensive modifications, working alongside people to perform the same everyday tasks. However, to function effectively in such settings, it must first be capable of executing the kinds of motions that humans can perform. In this context, success is measured less by proficiency in a single, task-specific behavior and more by the generality of performing a broad repertoire of coordinated, whole-body movements and skills with human-like fluidity. However, realizing human-level physical intelligence on humanoids presents significant challenges: their high degrees of freedom, intermittent and multi-contact interactions, and inevitable modeling mismatches demand control strategies that can capture the richness of human movement while withstanding real-world uncertainty. 
Here, we address this grand challenge with \emph{ZEST} (Zero-shot Embodied Skill Transfer), a unified framework for learning physical intelligence. \emph{ZEST} translates the diversity and expressiveness of human motion from heterogeneous data sources into a robust control policy. This approach enables a legged robot to execute a vast spectrum of skills zero-shot, without any per-task fine-tuning, circumventing the need for complex behavior discovery from scratch by leveraging human motion datasets as a source of general-purpose skills. 

Over the past decade, a popular approach for enabling diverse behaviors on legged robots has been to generate whole-body motion trajectories offline and then track them with a model-based controller. A prime example is Boston Dynamics’ Atlas humanoid, which has demonstrated impressive parkour and dance routines by leveraging references derived from keyframe animation. These motions are executed using a two-layer control architecture composed of a Model Predictive Control (MPC) online planner and an optimization-based whole-body controller \cite{Kuindersma2016AtlasWBC, BostonDynamics2024PickingUpMomentum, Kuindersma2020RecentProgressAtlas, AtlasGetsAGrip2025, HDAtlasManipulates}.
The research community has widely adopted the same model-based \emph{plan offline, track online} paradigm across a broad range of tasks for legged platforms. This includes humanoid robots using handrails to climb stairs \cite{KudrussOCP}, quadrupeds showing acrobatic behaviors \cite{li2024cafe}, retargeting motions from animals to robots with different morphologies \cite{RubenDOC}, and quadrupedal mobile manipulators tackling multi-contact tasks \cite{Sleiman2023MultiContact}.
Across these efforts, the core principle remains the same: use offline trajectory optimization---often in a multi-contact setting---to discover nontrivial, dynamically feasible behaviors that are difficult to design manually, and track these motions online with a general-purpose stabilizing controller.
This approach offers versatility and interpretability as a well-tuned controller can track a wide range of physically consistent motions without requiring skill-specific adjustments.
However, the resulting controllers tend to have stronger requirements on environment modeling, estimation, and reference fidelity, making them more difficult to deploy in scenarios involving high uncertainty or complex contact.

Recently, Reinforcement Learning (RL) has emerged as a powerful approach for control synthesis, providing exceptional agility and robustness. By learning to manage complex contact dynamics implicitly, RL policies can determine when and where to make contact in real-time, eliminating the need for predefined contact schedules and simplifying control design. Moreover, the ability to train at scale in high-fidelity simulators and transfer policies to hardware zero-shot \cite{RL_Nikita, RL_SimToRealSurvey} has accelerated progress in data-driven control \cite{RLLocomotionSurvey, RLHumanoidSurvey}. This success has led to state-of-the-art performance in legged locomotion, including robust navigation over challenging terrains in the wild \cite{RL_Joonho,Miki2022WildLocomotion,HumanoidPerceptiveLocomotion}, high-speed running \cite{AJSpot,HighSpeedRunning}, and extreme parkour \cite{Cheng2023ExtremeParkour,Hoeller2024AnymalParkour,Rudin2025ParkourWild}. 
Beyond locomotion, RL has also proven effective for dexterous manipulation tasks \cite{Andrychowicz2020DexterousManipulation, Handa2023Dextreme, HumanoidDexterousManipulation}, and whole-body mobile manipulation \cite{Ma2025Badminton,FALCON,DynamicBoxPushing,CuriosityRL}. For instance, it has enabled various applications in whole-body mobile manipulation, such as a quadruped to play badminton \cite{Ma2025Badminton}, a humanoid robot to perform forceful tasks such as carrying heavy loads, pulling carts, and opening spring-loaded doors \cite{FALCON}, and a quadrupedal manipulator to dynamically push and reorient large boxes \cite{DynamicBoxPushing}. Nonetheless, these \emph{tabula rasa} RL techniques remain sample-inefficient and highly sensitive to reward design; without careful shaping and regularization, learned policies can exploit the reward structure and exhibit unnatural or overly aggressive behaviors. 

To mitigate the limitations of \emph{tabula rasa} RL, an effective strategy, pioneered in computer graphics, is to use motion data as a prior to regularize policy learning. This data-guided recipe guides controllers toward natural movements, reducing the need for extensive reward engineering. A seminal example is \emph{DeepMimic} \cite{Peng2018DeepMimic}, which trained polices to imitate a collection of Motion Capture (MoCap) clips while relying on a single, consistent reward structure. This approach was later extended to enable user-steerable control with a motion matcher \cite{Bergamin2019DReCon}, achieve higher-fidelity tracking of complex motions \cite{Yuan2020RFC}, and scale imitation to massive motion libraries \cite{Luo2023PHC}.  
Adversarial imitation learning replaces direct imitation rewards with those inferred indirectly by matching the distribution of demonstrations. In Adversarial Motion Priors \cite{Peng2021AMP}, a discriminator trained to differentiate between state transitions from the dataset and those from policy rollouts provides a style reward that encourages natural motions. Subsequent variants introduce hierarchical or latent-skill formulations to mitigate mode collapse and improve reusability and steerability \cite{Peng2022ASE, Tessler2023CALM}.

More recent work closes the loop by coupling kinematic motion generators with physics-based controllers. For instance, text-driven diffusion models now generate plans for multi-task sequences \cite{Tevet2025CLoSD}, while other hierarchical approaches tackle agile terrain traversal by iteratively augmenting motion datasets \cite{Xu2025PARC}. This paradigm has also been extended to complex human-object interactions through methods like multi-teacher policy distillation \cite{Xu2025InterMimic}.
These works from computer graphics highlight a key theme: data---whether as explicit reference clips, adversarial priors, or learned skill embeddings---acts as a powerful regularizer that simplifies reward design, improves motion quality, and scales controllers to broad behavior repertoires.


\begin{figure*}[t!]
\captionsetup{format=plain}
    \centering
    \makebox[0pt]{\includegraphics[keepaspectratio, width = 
\textwidth]{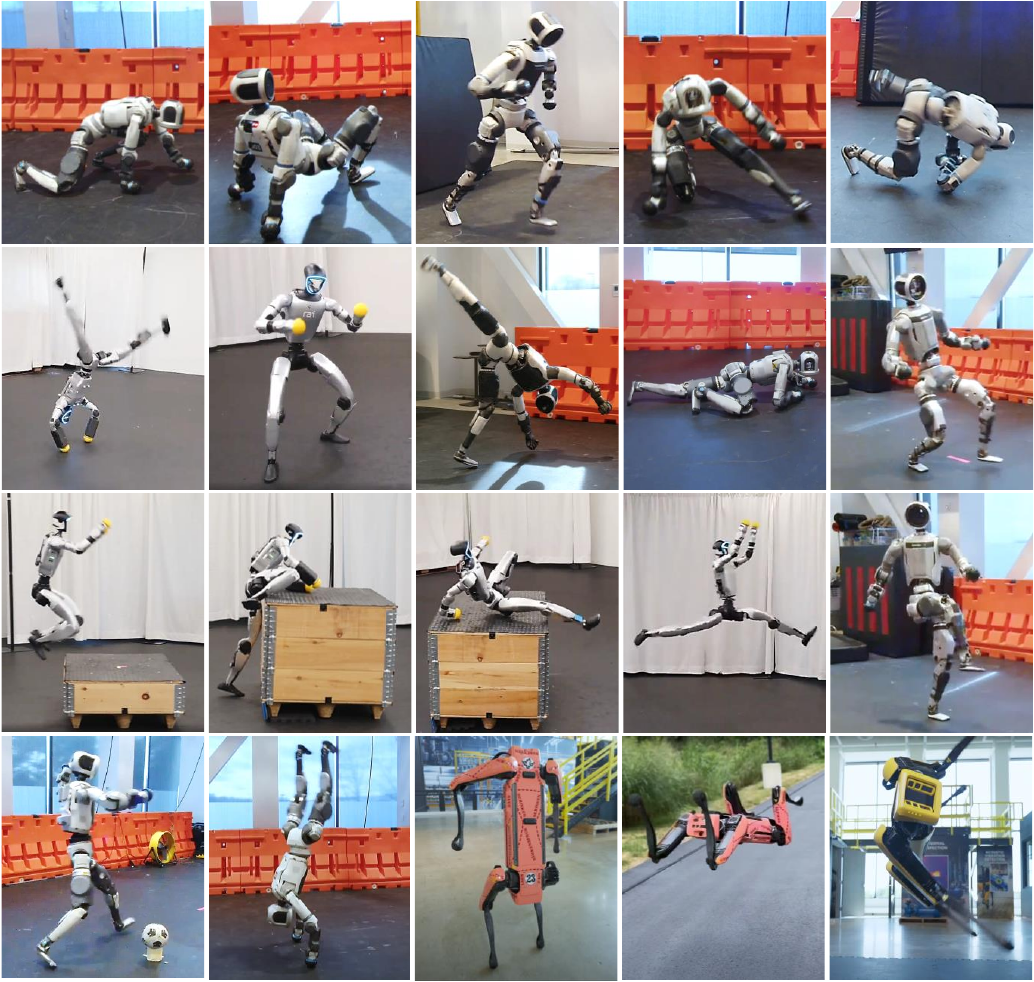}}
    \caption{\textbf{Hardware deployment of \emph{ZEST} across diverse data sources and robot morphologies.} In order of appearance from top left to bottom right, the figure illustrates the following behaviors. From MoCap: Crawl on all fours (Atlas), roll on all fours (Atlas), jog (Atlas), breakdance (Atlas), forward roll (Atlas), cartwheel (G1), table-tennis (G1), cartwheel (atlas), army crawl (Atlas). From ViCap: Dance snippet A (Atlas), jump onto box (G1), climb up/down box (G1), ballet (G1), dance snippet C, soccer kick (Atlas). From Animation: handstand invert (Atlas), handstand balance (Spot), continuous backflip (Spot), barrel roll (Spot).}
    \label{fig:motion_collage}
\end{figure*}

Despite strong results in simulation, deploying a unified tracking policy on hardware remains nontrivial. The sim-to-real gap---stemming from inevitable modeling discrepancies---is worsened by contact-rich behaviors and operations near the robot’s actuation limits: many dynamic or acrobatic motions become infeasible once real torque, speed, and safety constraints are enforced. In contrast, such limits are often ignored in computer graphics settings, allowing for aggressive exploration during training, which in turn facilitates learning. Policies must also operate under partial observability (e.g., missing global pose or root linear velocity), typically requiring either a state-estimation stack or a history-conditioned policy, both of which add complexity during training and deployment. 
Nevertheless, a growing body of work has begun to bridge this gap by leveraging real-to-sim strategies and various data sources. In \cite{Peng2020ImitateAnimals}, Peng et al. enable quadruped robots to perform diverse gaits by imitating real animal MoCap data. Grandia et al. \cite{Grandia2024BipedalCharacter} present an RL-based policy that executes expressive, real-time stage performances driven by animation engines. 
\emph{VideoMimic} \cite{Allshire2025VideoMimic} utilizes everyday videos to train a whole-body policy to execute contextual skills through a real-to-sim-to-real pipeline. \emph{ASAP} \cite{He2025ASAP} learns to perform agile whole-body skills by pre-training motion-tracking policies from human videos, collecting trajectories on hardware, and learning a ``delta-action'' model that reduces the sim-to-real gap. 
\emph{KungfuBot} \cite{KungfuBot} trains a policy for Unitree G1 to perform dynamic motions, such as acrobatic dance and kung-fu strikes, through multi-step motion processing (filtering, correction, retargeting with physical constraints). \emph{HuB} \cite{HUB} pushed the limits of the same hardware to achieve extreme balance in challenging one-legged stance tasks.

Using reference trajectories from model-based planners is another strategy for imitation-based RL. 
Methods like \emph{Opt-Mimic} \cite{Fuchioka2023OptMimic} and \emph{DTC} \cite{jenelten2024dtc} exemplify this approach by training policies to mimic optimized plans, thereby combining the precision of trajectory optimization with the robustness of an RL tracking controller.
Similarly, a two-stage RL framework can leverage a single optimal trajectory as initial guidance and then fine-tune purely for task completion, yielding robust locomotion policies \cite{Bogdanovic2022TOdemo}.
For multi-contact loco-manipulation, Sleiman et al. \cite{Sleiman2025GuidedRL} utilize a single offline demonstration to train policies for tasks like traversing spring-loaded doors and manipulating heavy dishwashers. The resulting policy generalizes beyond the reference, learning to handle object variability and discover novel recovery maneuvers.

Another line of work focuses on human-driven and generalist control, grounded in large-scale motion tracking and teacher–student distillation. \emph{OmniH2O} \cite{He2024OmniH2O} treats kinematic pose as a universal control interface and learns an autonomous whole-body policy by imitating a privileged teacher trained on large-scale retargeted/augmented datasets. \emph{HOVER} \cite{He2024HOVER} distills a motion imitation teacher into a student policy while applying proprioception and command masking with mode-specific and sparsity-based masks, resulting in a unified multi-mode command space for a whole-body tracking policy. Most recently, \emph{GMT} \cite{Chen2025GMT} combines an adaptive sampling strategy (biasing sampling towards harder-to-learn motions) with a Mixture-of-Experts teacher that enhances expressiveness and generalizability. Student policies distilled from this teacher track a broad spectrum of motions achieving state-of-the-art real-world performance.

Despite these advancements, a significant obstacle to scalability and robustness persists: the dependence on intricate or multi-stage pipelines that are often composed of handcrafted components. These complexities can include auxiliary shaping reward terms, contact labeling, multi-stage training, access to non-proprioceptive signals or to histories of proprioception, and per-behavior retuning. 
In this work, we present \emph{ZEST}, a unified, generic, and minimal recipe that trains policies in a single stage from heterogeneous motion data sources, including MoCap, Video-Captured (ViCap) motions, and keyframe animations. By design, this approach eliminates complex, domain-specific techniques such as contact labeling, extensive reward shaping, future-reference windows, observation histories, and state estimators, while avoiding multi-stage training pipelines. The resulting policy, implemented with a simple feedforward network, robustly tracks diverse motion references using a single set of hyperparameters for each robot and deploys zero-shot to hardware. 
We apply \emph{ZEST} to Boston Dynamics' electrically-actuated Atlas humanoid robot (approximately 1.8\,m and 100\,kg, comparable to an average adult human). The exact formulation and training recipe are applicable across robots with different sizes, weights, and embodiments, from the full-scale Atlas humanoid to the smaller Unitree G1 humanoid and the Spot quadruped. This versatility demonstrates a level of generality that represents a significant advance over specialized controllers tightly coupled to a single robot or behavior. 

Concurrent work reports closely related motion-imitation pipelines and transfer results on the G1 platform \cite{BeyondMimic,OmniRetarget,Sonic}. In this work, we evaluate a single, unified framework with a systematic, low-tuning training recipe across multiple robot platforms and multiple reference sources. This evaluation yields, to our knowledge, the first dynamic, multi-contact behaviors (e.g., army crawl, breakdance) on a full-size humanoid (Atlas) and the first video-to-behavior transfer of agile, contact-rich skills, such as box-climbing and dancing, to a physical humanoid (Unitree G1). Related work on ViCap-based sim-to-real transfer such as \emph{VideoMimic} \cite{Allshire2025VideoMimic} primarily focuses on locomotion-style behaviors, whereas our video-derived demonstrations emphasize complex dynamic multi-contact skills with intermittent whole-body contact events and sustained whole-body contact interactions.

Our policies utilize the next-step reference and current proprioceptive signals as observations, with only the previous action as the history. They output residual joint targets, which are then added to the reference and sent to a joint-level PD (Proportional-Derivative) controller.
To ground our simulation in reality, the PD gains are tuned using effective motor armature values derived from an analytical model that approximates closed-chain kinematics in actuators such as knees and ankles. For Spot, we incorporated more accurate models of its power system and actuators in simulation.
To handle long-horizon clips and scale beyond single skills, trajectories are split into fixed-duration bins, with difficulty level estimated via an Exponential Moving Average (EMA) of failure scores. A categorical sampler biases toward harder bins, facilitating long-horizon and multi-trajectory learning without catastrophic forgetting.
To help stabilize training and avoid long convergence times for highly dynamic behaviors---especially those involving large base rotations---a virtual assistive wrench is computed in a model-based fashion and applied directly to the robot's base. Its magnitude is adapted based on the per-bin difficulty levels and eventually decays to zero as tracking performance improves, eliminating the need for a handcrafted curriculum. Training is performed entirely in simulation with moderate domain randomization (e.g., impulsive pushes, sensor noise, and variations in friction coefficients and masses). 
The result of this framework is a pipeline that demonstrates remarkable robustness from data to deployment. It effectively processes reference motions from diverse sources, successfully handling high-fidelity MoCap data, noisy ViCap data with artifacts like pose jitter and foot-skidding, and kinematically clean but often physically infeasible animation data. The framework's ability to retarget motions from all three sources highlights its resilience to a wide range of data imperfections. This entire training process results in policies that are deployed to the hardware zero-shot via an automated pipeline. 

\begin{figure}[t]
    \centering
  \makebox[0pt]{\href{https://youtu.be/CXyIXiT4zhA}{\includegraphics[width=\linewidth, keepaspectratio]{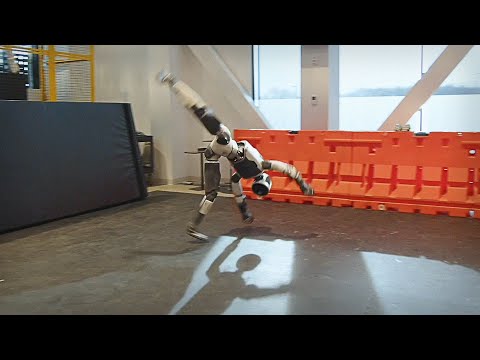}}}
\captionsetup{type=Movie}
\caption{\label{fig:plate} \textbf{Summary of the \emph{ZEST} framework and its hardware results.}}
\end{figure}


\section*{RESULTS}

\noindent Movie~1 summarizes the methodology and results of the presented work. In this section, we outline our evaluation procedure and present our empirical findings. Unless otherwise indicated, all presented results are obtained on hardware. Each experiment uses a specialized policy trained solely on the target skill. This specialization avoids the extended training required for a purely multi-skill model to sample the target reference as often as in the single-skill setting and, in practice, delivers stronger and more consistent tracking performance within a fixed time budget. Therefore, we do not deploy a multi-skill policy on hardware, but only use it to perform simulation-based analyses. Each policy was trained for approximately 10 hours (about 7k iterations) on a single NVIDIA L4 GPU and validated across multiple independent hardware trials with consistent success. Evaluations span multiple embodiments---Boston Dynamics' electrically actuated Atlas humanoid, Unitree’s G1 humanoid, and Boston Dynamics' Spot quadruped---and are organized by reference source: MoCap, ViCap, and keyframe animation. Figure~\ref{fig:motion_collage} shows representative frame snapshots across all skills, per-skill clips are provided in Movie~1, and the corresponding reference motions are included in Movies~S1--S3. Table~\ref{tab:all_motions_eval} lists all motions, and for each, marks whether it involves multiple body contacts with the ground (i.e., apart from the robot's feet), and reports tracking metrics for joint angles, base orientation, and base angular velocity. The base orientation is represented by its roll and pitch components. We exclude yaw, as it is not directly observable from the on-board IMU and is subject to significant drift. Next, we evaluate a multi-skill policy in simulation and use it to justify our key design choices via simulation-based ablation studies. Finally, we include a simulation-based comparison with a state-of-the-art whole-body MPC baseline developed at Boston Dynamics. 


\subsection*{MoCap-Derived Human-Like Skills on Hardware}

\noindent MoCap references enable a broad range of behaviors on Atlas (Table~\ref{tab:all_motions_eval}; Movie~S1). As shown in Movie~S1, we deploy \emph{walk}, \emph{jog}, \emph{forward roll}, \emph{roll on all fours}, \emph{crawl on all fours}, \emph{army crawl}, \emph{cartwheel}, and a short \emph{breakdance} on hardware, and we also evaluate a MoCap \emph{table-tennis} and a \emph{cartwheel} motion on Unitree’s G1 using the same formulation (with actuator parameters adapted to that platform).

We highlight salient aspects of these MoCap-driven behaviors. For \emph{walk} and \emph{jog}, foot--ground interactions are straightforward, yet several natural-gait hallmarks emerge: a clear heel--toe rollover rather than flat-foot placement; near full-knee extension despite proximity to joint limits; and graceful, quiet transitions---both within gait cycles and when decelerating from a fast jog to stance---without stomping, chatter, or audible impacts. These features are extremely challenging to induce through pure reward shaping alone. We attribute their emergence to high-fidelity MoCap references that preserve human nuance, actuator models tuned to be responsive without excessive stiffness, and regularization that discourages impulsive transients.

The same formulation also accommodates complex multi-contact behaviors without prescribing contact schedules or constraining base height or orientation to upright configurations. All motions begin from a standing configuration and return to standing at completion. \emph{Forward roll} throws the body forward from stance, makes initial forearm contact, and completes a full forward roll with back contact. \emph{Roll on all fours} and \emph{crawl on all fours} employ intermittent heel--toe, knee, and wrist contacts; the former executes a lateral roll, while the latter advances forward under low ground clearance. \emph{Army crawl} maintains near-continuous arm--leg--torso contact with deliberate sliding to translate forward. \emph{Cartwheel} alternates hand--foot supports and passes through transient hand-supported inversions that require accurate hand placement and high angular-rate control. The most challenging case is the \emph{breakdance} sequence, which combines extended body--ground contact phases with rapid leg sweeps and includes segments operated near joint limits where a leg---with its knee fully extended---threads beneath the opposite leg, inducing unavoidable self-contact.

\begin{table*}[t!]
\centering
\caption{\textbf{Evaluation across motions, grouped by reference source.} A skill is labeled as \emph{multi-contact} if any body part other than the feet interacts with the ground. Metrics are reported as \emph{mean absolute error (MAE), mean angular distance (MAD), and mean L2 error (ML2)} over successful trials; lower is better. MAE$(\bm q)$ is for joint-position error; MAD$(R)$ is for base orientation error; ML2$(\boldsymbol\omega)$ is for base angular velocity errors, respectively. We also include data for the maximum errors in joint position and orientation, denoted as max$(\bm q)$ and max$(R)$ respectively.}
\label{tab:all_motions_eval}
\renewcommand\theadfont{\footnotesize\bfseries}

\begingroup
\sisetup{round-mode=places, round-precision=3} 
\setlength{\tabcolsep}{4pt}
\renewcommand{\arraystretch}{0.65}
\begin{tabular*}{\textwidth}{@{\extracolsep{\fill}}
  l l c
  S[table-format=2.2, round-precision=2] 
  S[table-format=1.3] 
  S[table-format=1.3] 
  S[table-format=1.3] 
  S[table-format=1.3] 
  S[table-format=1.3] 
  @{}}
\toprule
\multicolumn{1}{l}{\thead{Motion}} &
\multicolumn{1}{l}{\thead{Robot}} &
\multicolumn{1}{c}{\thead{Multi-Contact\\Skill (\xmark/\checkmark)}} &
\multicolumn{1}{c}{\thead{Duration\\(s)}} &
\multicolumn{1}{c}{\thead{MAE $(\bm q)$\\(rad)}} &
\multicolumn{1}{c}{\thead{MAD $(R)$\\(rad)}} &
\multicolumn{1}{c}{\thead{ML2 $(\boldsymbol\omega)$\\(rad/s)}} &
\multicolumn{1}{c}{\thead{max $(\bm q)$\\(rad)}} &
\multicolumn{1}{c}{\thead{max $(R)$\\(rad)}}\\
\midrule

\multicolumn{9}{@{}l}{\textbf{MoCap-Derived Motions}} \\
\addlinespace[0.3em]
Walk                 & Atlas & \xmark     & 8.47  & 0.056848 & 0.030011  & 0.364151 & 0.486572 & 0.078402 \\
Jog                  & Atlas & \xmark     & 5.89  & 0.041299 & 0.059295  & 0.699386 & 0.642535 & 0.165264 \\
Forward roll         & Atlas & \checkmark & 7.11  & 0.061586 & 0.137010  & 1.654138 & 0.907468 & 0.527964 \\
Roll on all fours    & Atlas & \checkmark & 7.80  & 0.063823 & 0.080306  & 1.604383 & 0.642819 & 0.202891 \\
Crawl on all fours    & Atlas & \checkmark & 14.74  & 0.055822 & 0.096739  & 0.921553 & 0.764109 & 0.239705 \\
Army crawl            & Atlas & \checkmark & 16.66  & 0.103255 & 0.075445  & 1.010727 & 0.638430 & 0.276292 \\
Cartwheel             & Atlas & \checkmark & 5.98  & 0.083514 & 0.064348  & 1.733776 & 0.672884 & 0.234598 \\
Breakdance   & Atlas & \checkmark & 6.56  & 0.079027 & 0.133372  & 1.911055 & 0.825025 & 0.364705 \\
Cartwheel    & G1    & \checkmark &  8.0   & 0.078 & 0.331  & 0.886 & 0.660 & 0.892 \\
Table-tennis  & G1    & \xmark     & 29.98 & 0.108    & 0.402     & 1.827    & 2.483    & 0.637 \\
\addlinespace[0.5em]
\midrule

\multicolumn{9}{@{}l}{\textbf{ViCap-Derived Motions}} \\
\addlinespace[0.3em]
Soccer kick           & Atlas & \xmark     & 6.00  & 0.049252 & 0.050408  & 1.253406 & 0.653306 & 0.140620 \\
Dance snippet A       & Atlas & \xmark     & 12.70  & 0.050629 & 0.038604  & 0.414272 & 0.769049 & 0.079681 \\
Dance snippet B       & Atlas & \xmark     & 9.40  & 0.054249 & 0.065766  & 1.463322 & 0.761656 & 0.238778 \\
Dance snippet C       & Atlas & \xmark     & 11.59  &  0.054567 & 0.046293  & 1.248154 & 0.759036 & 0.125329 \\
Ballet sequence       & G1      & \xmark     & 7.00   & 0.061 & 0.186  & 1.395 & 1.101 & 0.588 \\
Jump onto box         & G1      & \xmark     & 5.00   & 0.041 & 0.259  & 0.936 & 0.883 & 0.914 \\
Climb up box          & G1      & \checkmark & 9.18   & 0.073 & 0.385  & 0.855 & 1.103 & 0.798 \\
Climb down box        & G1      & \checkmark & 10.62  & 0.108 & 0.394  & 0.836 & 0.957 & 0.867 \\
\addlinespace[0.5em]
\midrule

\multicolumn{9}{@{}l}{\textbf{Animation-Derived Motions}} \\
\addlinespace[0.3em]
Handstand invert & Atlas & \checkmark & 5.18  & 0.073904 & 0.082358  &  0.465313 &  0.949880 &  0.449680 \\
Handstand balance     & Spot    & \xmark & 4.30  & 0.0753726072316811 & 0.063911  & 0.408573 & 0.531845 & 0.235228 \\
Continuous backflip              & Spot    & \xmark     & 6.00  & 0.192162 & 0.149614  & 0.732049 & 2.381236 & 0.528300 \\
Barrel roll*           & Spot    & \xmark & 2.45  & 0.145794 & 0.839707*  & 1.450854 &  1.067999 & 1.570796 \\
Happy dog             & Spot    & \xmark     & 12.00  & 0.10668668926816251 & 0.069333  & 1.008193 & 1.180046 & 0.348501 \\
\addlinespace[0.5em]
\midrule
\multicolumn{9}{l}{\footnotesize \textsuperscript{*} During barrel roll, the Spot IMU saturated mid-flight; therefore, IMU data were excluded from that segment.} \\
\bottomrule
\end{tabular*}
\endgroup
\end{table*}

These motions are difficult precisely because they invoke contacts beyond the feet (hands, knees, forearms, torso, back), placing them outside the design envelope of standard whole-body control pipelines. For model-based frameworks, the intricacy of modeling contact phenomena---high-impulse events, stick--slip transitions, and extended body--ground interactions over complex geometries---poses a major obstacle to designing stabilizing controllers for multi-contact behaviors. Accordingly, these methods typically predefine the active end-effectors and avoid motions with whole-body contacts. Even for RL policies trained in simulation, these regimes magnify the sim-to-real gap, as small mismatches in contact, friction, compliance, and geometry compound over extended contact-rich phases. Consistent with Table~\ref{tab:all_motions_eval}, simpler foot-only gaits (\emph{walk}, \emph{jog}) exhibit lower tracking errors across $\mathrm{MAE}(\bm q)$ and $\mathrm{MAD}(R)$, whereas more dynamic, multi-contact skills (\emph{forward roll}, \emph{roll on all fours}, \emph{crawl on all fours}, \emph{army crawl}, \emph{cartwheel}, \emph{breakdance}) show higher orientation and angular-velocity discrepancies.

Finally, we also imitate a \emph{cartwheel} motion and a \emph{table-tennis} on Unitree’s G1. While our G1 evaluations are predominantly ViCap-derived motions (next section), this result provides an additional check of humanoid-to-humanoid cross-morphology transfer while holding the formulation constant across varying data sources. The G1's \emph{Cartwheel} reference, although approximately 2s longer than the version for Atlas, demonstrates a comparable motion quality. However, it has a higher orientation error, which we attribute to differences in the quality of the IMUs between the two platforms. Another challenging long-horizon task is the \emph{table-tennis}, which requires coordinating upper-body swings with dynamic base maneuvers. Its extended duration makes it difficult to learn, and our adaptive sampling strategy was crucial in achieving a successful outcome for this behavior. 

\subsection*{ViCap-Derived Human-Like Skills on Hardware}

\noindent Our experiments with ViCap references highlight robustness to noisy, video-reconstructed motions while maintaining fluid, temporally coherent motion over long sequences (Movie~S2). We employ video demonstrations captured in the wild with a handheld phone camera, yielding a fast and practical data generation pipeline: \emph{record $\rightarrow$ reconstruct $\rightarrow$ retarget $\rightarrow$ train $\rightarrow$ execute}, where \emph{reconstruct} denotes 3D pose estimation with temporal smoothing. Converting a recorded video into a retargeted reference motion takes minutes, so in practice, we can capture a new clip in the morning, train during the day, and run it on hardware that evening.

On Atlas, we execute a \emph{soccer kick} motion and three \emph{dance} snippets (A--C). These sequences include extended single-foot support phases and multiple consecutive hops, requiring sustained balance with precise foot placement and timing. Despite visible artifacts in the references (pose jitter and foot skidding; Movie~S2), we still achieve zero-shot sim-to-real transfer while preserving expressivity, without heavy motion post-processing (e.g., physics-aware trajectory optimization for high-fidelity dynamic retargeting); as reported in Table~\ref{tab:all_motions_eval}, tracking errors remain surprisingly low (e.g., $\mathrm{MAE}(\bm q)$ and $\mathrm{MAD}(R)$ near $0.05$\,rad). We attribute this to a reward formulation that does not force tight adherence to the reference root trajectory: the policy is encouraged to track the global motion in a relaxed fashion, rejecting obvious artifacts while still solving the task. Moreover, our MDP and rewards do not require contact labels, so the policy is agnostic to noisy contact timings/schedules extracted from the ViCap reference.

We further demonstrate cross-embodiment transfer on Unitree’s G1 with an athletic \emph{ballet sequence} featuring a sustained aerial phase with both legs abducted in mid-air and extended single-foot support near the edge of stability, stressing balance, orientation regulation, and precise foot placement. We also evaluate scene-interactive \emph{box} skills on G1: \emph{jump onto box}, \emph{climb up box}, and \emph{climb down box}. These experiments involve whole-body contacts and push the system near its operational limits. 
Instead of providing the policy with the box's precise location through perception or MoCap, we test its ability to execute these skills robustly using a minimalist, proprioceptive-only interface. To achieve this, we train the policy with slight randomizations in the initial relative pose between the robot and the box. At test time, the policy reliably handles this variability without additional sensing. 
Furthermore, these scene-interactive skills reflect a combination of challenges, including (i) compounded modeling mismatches arising from multi-contact interactions, (ii) discrepancies between the simulated box and the physical platform (notably compliance and friction), (iii) lack of box--robot relative-positioning feedback, and (iv) artifacts in the reconstructed references. Despite these factors, hardware executions are completed reliably, and the characteristic style of the source motions remains evident. As highlighted in Table~\ref{tab:all_motions_eval}, errors remain reasonable overall and are comparable to those observed for the more challenging behaviors presented in the MoCap section. To further characterize repeatability and robustness on these challenging behaviors, we conduct additional variability testing. In Movie~S6, we execute box climb up and box climb down five times consecutively, achieving five out of five successful runs under conditions kept as close as possible to the nominal training setup (i.e., the box height set to its nominal value of 0.75\,m and the robot initialized near the nominal pose with respect to the box). In Movie~S7, we stress-test the policy's robustness by varying the initial robot--box relative pose around the nominal configuration. Since these policies are inherently invariant to lateral offsets relative to the box center, we randomize the robot's initial offset in \(x\) and yaw while keeping \(y\) close to the nominal value. All trials initialized within the training distribution succeed (\(\pm 10\)~cm in \(x\) and \(\pm 0.3\)~rad in yaw), as highlighted in Figure~S5. Beyond this range, we observe occasional successes, while failures occur for initial states outside the training distribution. We additionally test robustness to mass variability by attaching a 2\,kg payload to the torso and still obtain successful executions for both box climb up and box climb down. Finally, we vary the box height beyond the randomized training range. Although training uses a nominal height of 0.75\,m with randomization over \([0.70, 0.80]\)~m, box climb up executes successfully at 0.55\,m, while box climb down succeeds at both 0.55\,m and 0.95\,m.

\subsection*{Keyframe-Animated Skills for Humanoids and Quadrupeds on Hardware}



\noindent Keyframe animation broadens our motion library beyond natural human movement. For humanoids such as Atlas, it enables us to exploit robot-specific degrees of freedom not represented in human motion capture, such as continuous joints in the arms, legs, or spine. These joints enable configurations that human anatomy cannot achieve. For instance, we achieve a \emph{handstand invert} motion by rotating the robot's back--yaw joint while maintaining stable hand support (Movie~S3).

Keyframe animation is equally important for platforms that lack a human form, where collecting human-like demonstrations is inherently difficult. To assess generality under a larger morphology gap, we apply the same framework on Boston Dynamics' Spot, where we demonstrate a \emph{handstand balance} (forelegs), a \emph{continuous backflip}, a \emph{barrel roll}, and a playful \emph{happy dog} motion.

Although animation references are kinematically clean and temporally coherent, several sequences are dynamically aggressive or partially infeasible for the physical system. In these cases, the learned policy intentionally relaxes adherence to the reference---especially for base orientation and angular-rate profiles---to remain feasible, which increases reported means while preserving overall execution quality. This pattern is visible in Table~\ref{tab:all_motions_eval} when contrasting $\mathrm{MAE}(\bm q)$ and $\mathrm{MAD}(R)$ with the corresponding maxima: for Atlas \emph{handstand invert}, the means are moderate while $\max(\bm q)$ and $\max(R)$ spike during transient inversions; on Spot, \emph{backflip} and \emph{barrel roll} show larger orientation and angular-velocity errors (see $\mathrm{MAD}(R)$ and $\mathrm{ML2}(\boldsymbol\omega)$) and high maxima, reflecting brief deviations from the reference for the sake of feasibility and stability. 
Notably, the policy's robustness is highlighted on the Spot robot for the \emph{barrel roll} motion, where it completes the motion even after the on-board IMU sensor saturates mid-maneuver. 

\begin{figure*}[t!]
    \centering
    \begin{subfigure}[b]{0.24\textwidth}
        \centering
        \includegraphics[width=\textwidth]{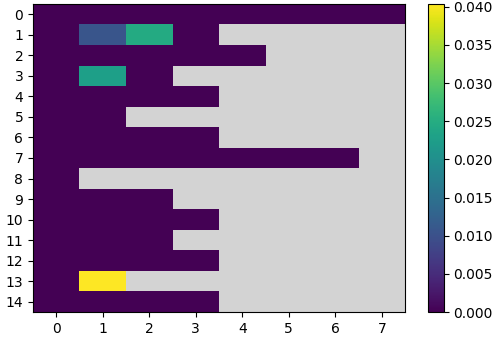}
        \caption{Wrench scaling.}
        \label{fig:wrench_scale}
    \end{subfigure}
    \hfill
    \begin{subfigure}[b]{0.24\textwidth}
        \centering
        \includegraphics[width=\textwidth]{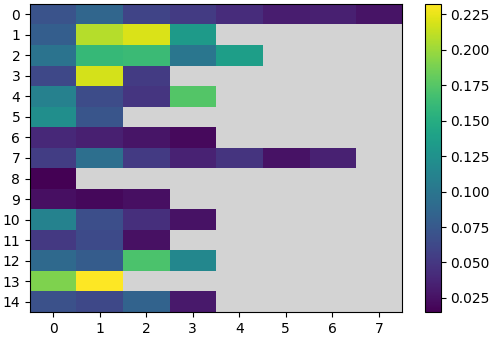}
        \caption{Failure levels.}
        \label{fig:failure_level}
    \end{subfigure}
    \hfill
    \begin{subfigure}[b]{0.24\textwidth}
        \centering
        \includegraphics[width=\textwidth]{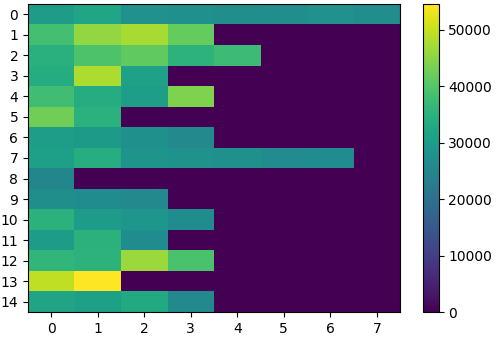}
        \caption{Trajectory-bin counts.}
        \label{fig:traj_bin_count}
    \end{subfigure}
    \hfill
    \begin{subfigure}[b]{0.24\textwidth}
        \centering
        \includegraphics[width=\textwidth]{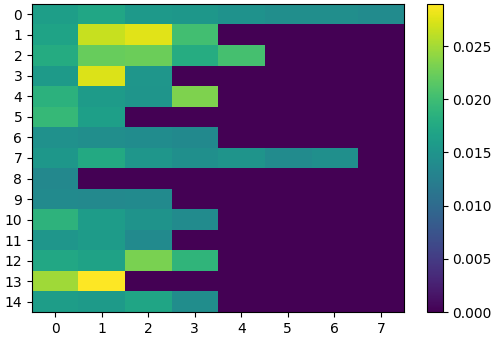}
        \caption{Sampling probabilities.}
        \label{fig:sampling_probs}
    \end{subfigure}

    \vskip\baselineskip

    \begin{subfigure}[b]{0.48\textwidth}
        \centering
        \includegraphics[width=\textwidth]{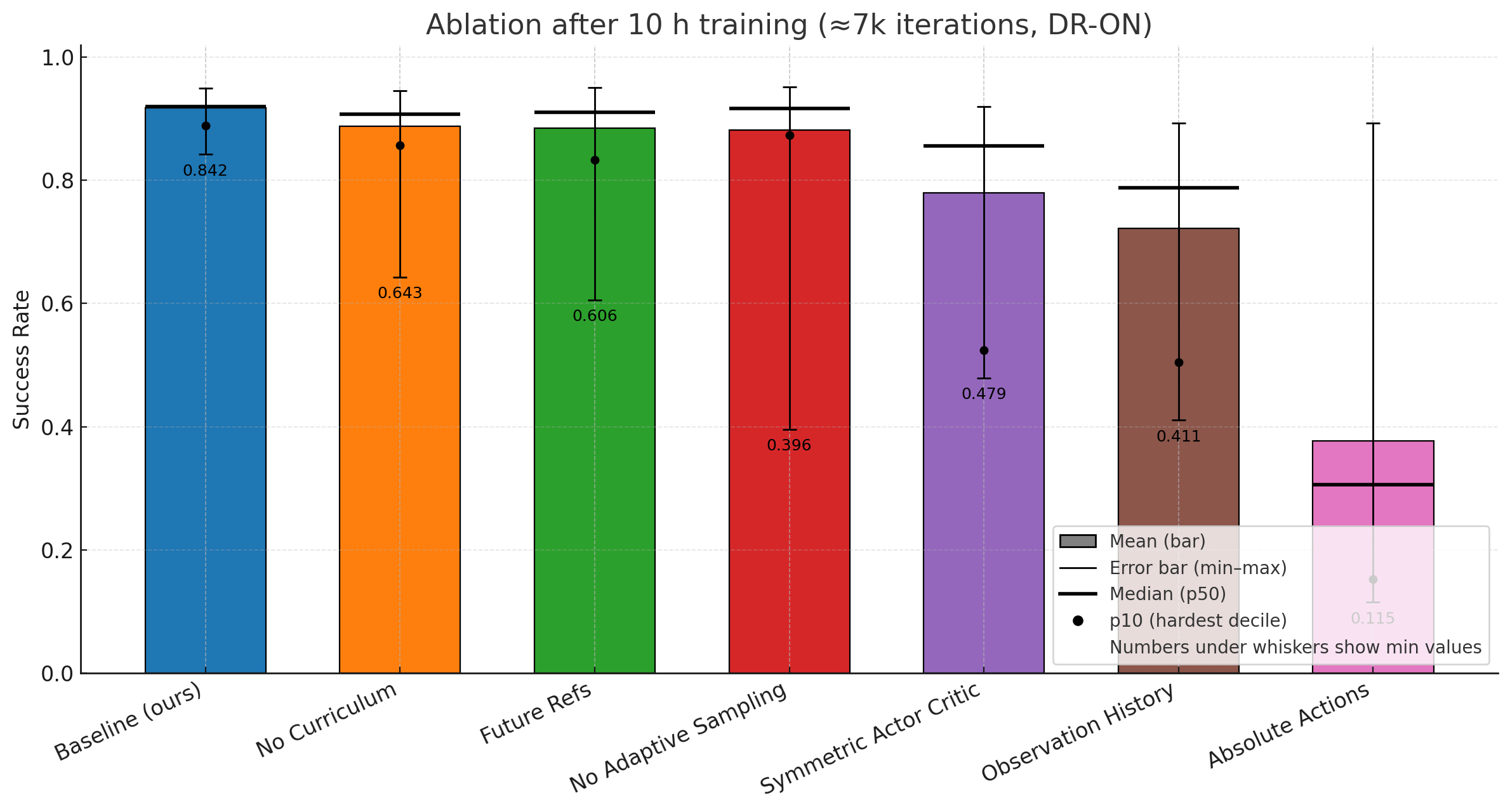}
        \caption{Ablation after 10 h training ($\approx$7k iterations).}
        \label{fig:ablation_10h}
    \end{subfigure}
    \hfill
    \begin{subfigure}[b]{0.48\textwidth}
        \centering
        \includegraphics[width=\textwidth]{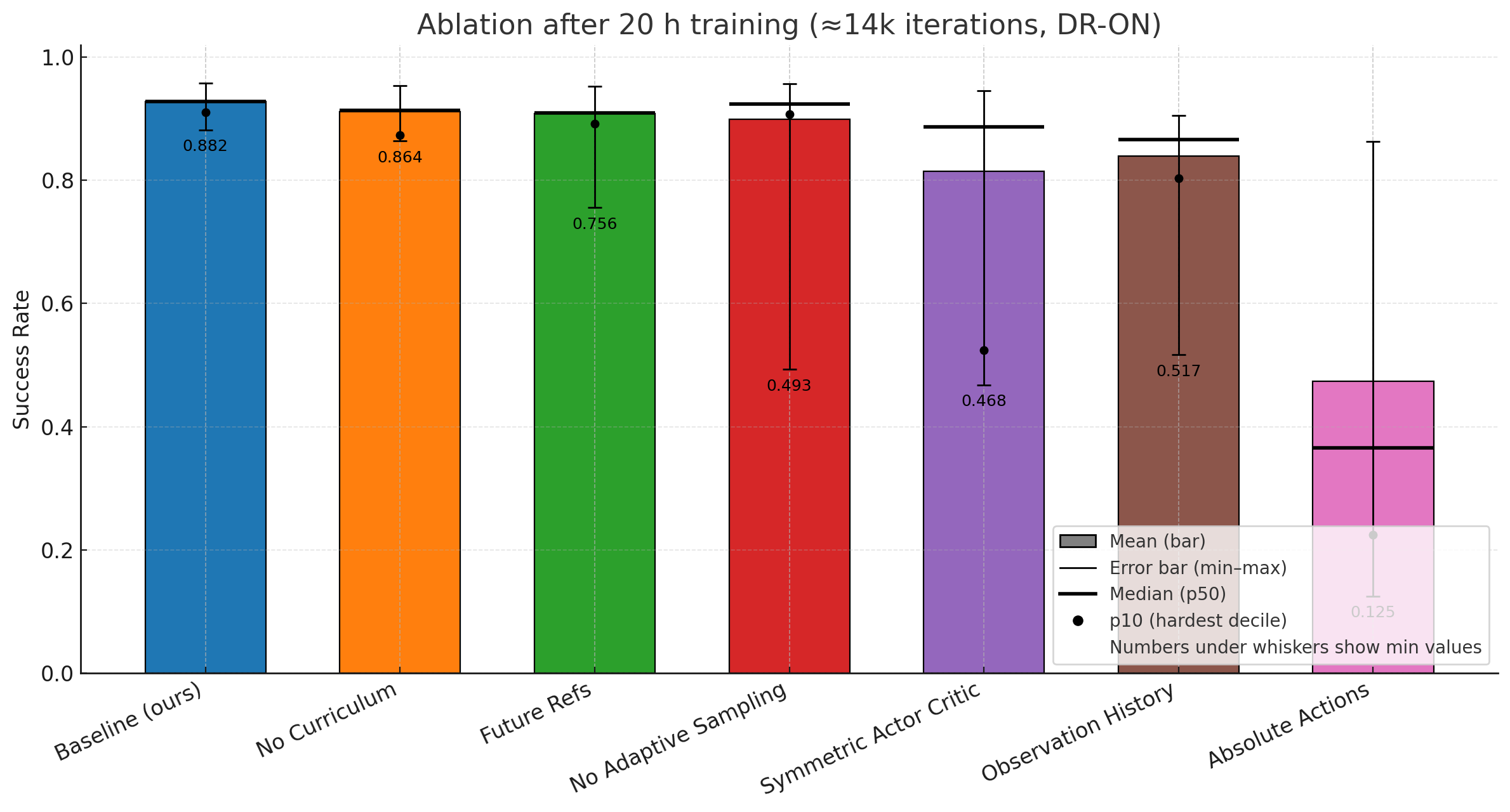}
        \caption{Ablation after 20 h training ($\approx$14k iterations).}
        \label{fig:ablation_20h}
    \end{subfigure}

    \caption{
        \textbf{Simulation-based evaluation and ablation studies.}
        (\textbf{a}–\textbf{d}) A mid-training snapshot of the signals driving our adaptive curriculum: (\textbf{a}) assistive wrench scaling, which is modulated by failure, (\textbf{b}) per-bin failure levels, (\textbf{c}) bin visitation counts, and (\textbf{d}) sampling probabilities. Higher failure rates increase the sampling probability, and over time, visitation counts rise where failures are frequent.
        (\textbf{e}–\textbf{f}) Success rates for the baseline and ablations after 10~h and 20~h of training under complete domain randomization. Bars show mean success; whiskers show min–max; internal lines mark median; dots denote p10; numbers under lower whiskers report the minimum. The plots show that the assistive curriculum and adaptive sampling are critical for performance and sample efficiency. Removing privileged information from the critic or using absolute actions significantly degrades robustness, while adding observation/reference windows hinders learning.
    }
    \label{fig:ablation_and_adaptive_sampling}
\end{figure*}

\subsection*{Simulation-Based Evaluation and Ablation Studies}
\noindent We train a single multi-skill policy for Atlas over 15 motions with diverse lengths and difficulties drawn from MoCap, ViCap, and keyframe animations, ranging from simple gaits (walking, squatting) to multi-contact and acrobatic sequences (army crawl, breakdancing, cartwheel-to-backflip). To inspect how the method operates during learning, Figure~\ref{fig:ablation_and_adaptive_sampling}(a–d) shows a representative mid-training snapshot of our adaptive sampling signals: failure levels per trajectory bin, the resulting sampling probabilities, the number of times each bin has been visited, and the assistive-wrench scaling applied when starting from that segment. Higher failure rates increase sampling probability and, over time, visitation counts rise where failures are frequent. The easiest bin is the animated squat (trajectory index~8), while the cartwheel-to-backflip sequence (trajectory index~13) is the hardest; the sampler thus allocates more training to the latter and less to the former, while modulating the assistive wrench accordingly.

Evaluation proceeds by initializing each episode at the start of a randomly sampled reference trajectory and rolling it to completion. For each reference, we run 10000 rollouts under full domain randomization, sampling external pushes, observation noise, robot link masses, friction coefficients, and perturbations of the initial state. We quantify performance and robustness via ablations presented in Figure~\ref{fig:ablation_and_adaptive_sampling}(e–f), which report success rates after 10\,h ($\approx$7k iterations) and 20\,h ($\approx$14k iterations) of training. At 10\,h, our method achieves the highest mean success and markedly better lower-tail performance across behaviors. By 20\,h, several ablations narrow the mean gap, yet the baseline remains more robust, with higher p10 and minimum success values that lie closer to the mean.

We ablate, one at a time, curriculum, adaptive sampling, reference and observation windows, action space, and actor–critic observation design. Removing curriculum still converges, but its 20\,h performance roughly matches our method at 10\,h, indicating that a minimal variant without curriculum is feasible but less sample-efficient, particularly on motions with large base-orientation variability (e.g., trajectory index~13). Removing adaptive sampling decreases performance because the policy undersamples hard bins. The heatmaps support this mechanism, as sampling probabilities track failure levels and concentrate updates where failure is high, which the ablation cannot reproduce. 

To better contextualize the ablations over the wrench-assist curriculum and the adaptive sampling strategy, we summarize the roles played by each component. The assistive wrench primarily serves as an exploration and stabilization aid for highly dynamic motions (e.g., the cartwheel-to-backflip sequence) that frequently terminate early in training; in these regimes, it reduces catastrophic failures, stabilizes learning, and typically improves sample efficiency when reset-based exploration (i.e., Reference State Initialization) alone is insufficient. For simpler motions with fewer early failures, we find the wrench can often be reduced or removed with limited impact. Adaptive sampling is most beneficial when training on long-horizon trajectories or large motion libraries, and when the goal is to track each phase of each motion well rather than optimizing average performance. Without adaptive sampling, training can under-emphasize rare or difficult trajectory segments, leading to persistent weaknesses on those segments and degraded lower-tail performance.

Using longer reference or observation windows---either 20 steps of history or 20 steps of future references (0.4\,s each)---increases input dimensionality and redundancy, complicates credit assignment, and slows or prevents convergence under the same budget. This negative result may be specific to our hyperparameters and architecture; sequence-to-sequence models or different tuning could yield different conclusions, but we did not pursue such changes to keep the setup minimal and focused. We provide a finer-grained sweep over future-reference horizons and observation-history lengths (5, 10, 15, and 20 steps) in the Supplementary Materials (Figure S3), which reveals a gradual degradation in performance as the window length increases.

Matching actor and critic inputs by stripping privileged observations from the critic degrades value estimation. The critic benefits from privileged signals such as the full root state, contact forces, key-body positions, and the assistive-wrench signal, which reflects curriculum-induced dynamics. Without these, value estimates become noisier, leading to weaker policy updates. Finally, absolute actions (as opposed to residual actions) yield the worst performance. Because episodes are initialized on the reference throughout training, small policy outputs---which typically occur during early stages of training---can induce large PD tracking errors and hence large joint torques, destabilizing rollouts. In addition, absolute actions impose a larger exploration burden: the policy must synthesize the full joint commands, whereas residual actions leverage the feedforward reference and learn only the required offsets, leading to faster and more stable learning.

Additional details on the reference motions used for ablation are provided in the Supplementary Materials (Table~S1), along with a video showcasing evaluation rollouts from the multi-skill policy (Movie~S4). We further report simulation-based robustness sweeps across domain-randomization factors in the Supplementary Materials (Figure S4).

\begin{table}[t]
\centering
\caption{\textbf{Simulation-based comparison of MPC vs.\ a multi-skill RL policy.} The RL policy is trained on all Atlas motions in Table~\ref{tab:all_motions_eval}, but evaluations are performed over a subset of these motions. A dash (--) indicates failure to execute the behavior.}
\label{tab:mpc_vs_rl_atlas_subset_onecol}
\setlength{\tabcolsep}{6pt}           
\renewcommand{\arraystretch}{1.15}    
\small                                
\newcommand{\na}{\multicolumn{1}{c}{--}}
\begin{tabular}{@{}l
                S[table-format=1.6] S[table-format=1.6]
                S[table-format=1.6] S[table-format=1.6]
                @{}}
\toprule
\multicolumn{1}{c}{\textbf{Motion}} &
\multicolumn{2}{c}{$\mathrm{MAE}(\bm q)$ (rad)} &
\multicolumn{2}{c}{$\mathrm{MAD}(R)$ (rad)} \\
\cmidrule(lr){2-3}\cmidrule(lr){4-5}
& \multicolumn{1}{c}{MPC} & \multicolumn{1}{c}{RL}
& \multicolumn{1}{c}{MPC} & \multicolumn{1}{c}{RL} \\
\midrule
Walk               & 0.046695 & 0.055393 & 0.020235 & 0.027978 \\
Jog                & 0.117239 & 0.075721 & 0.109940 & 0.062128 \\
Handstand invert   & \na      & 0.146578 & \na      & 0.124962 \\
Cartwheel          & 0.237148 & 0.088191 & 0.059717 & 0.062788 \\
Roll on all fours  & \na      & 0.086242 & \na      & 0.091448 \\
Dance snippet B    & \na      & 0.096435 & \na      & 0.150253 \\
\bottomrule
\end{tabular}
\end{table}

\subsection*{Simulation-Based Benchmarking Against MPC}
\noindent We compare a multi-skill RL policy trained jointly on all Atlas motions presented in Table~\ref{tab:all_motions_eval} against a whole-body MPC baseline developed at Boston Dynamics, which is representative of state-of-the-art model-based controllers \cite{BostonDynamics2024PickingUpMomentum}. Evaluations use a subset of motions drawn from all three reference sources (MoCap, ViCap, and animation). For MPC, we extract contact schedules from each reference using a simple heuristic that considers the distance between the foot and the ground, as well as the foot's velocity. This heuristic is relatively straightforward for animated sequences (e.g., \emph{handstand invert}), but it becomes more error-prone for MoCap and particularly ViCap motions, where unrealistic foot sliding and noisy contact timings lead to mislabelled contacts and necessitate careful manual tuning. The MPC stack natively supports interaction with feet and hands, but not contacts beyond those end-effectors. As a result, motions that involve knee, torso, or forearm contacts (e.g., \emph{army crawl}, \emph{forward roll}, \emph{crawl on all fours}) fall outside its default capability and were excluded from the MPC comparison. In contrast, the learned policy does not require explicit contact labeling, is more tolerant of model mismatch and contact-timing errors, and readily accommodates interactions with arbitrary body parts. 

For a quantitative comparison, Table~\ref{tab:mpc_vs_rl_atlas_subset_onecol} reports $\mathrm{MAE}(\bm q)$ and $\mathrm{MAD}(R)$ for MPC and RL.
Empirically, for \emph{walk}, MPC and RL perform nearly the same: the contact schedule is clean and easy to label, and the trajectory never pushes the robot beyond its operational bounds, which suits MPC. On the other hand, RL outperforms MPC on the more dynamic behaviors, namely \emph{cartwheel} and \emph{jog}. In particular, for \emph{jog}, we observe MPC degradation tied to inconsistent contact annotation at the start of the reference despite extensive manual refinements. Several sequences fail under MPC (reported as ``--'' in the MPC columns): \emph{dance B} (inaccurate contact sequence and sliding), \emph{handstand invert} (controller saturation on arm torque under a point-contact constraint), and \emph{roll on all fours} (difficulty regulating extended toe-contact phases without knee collisions). The videos of the references augmented with contact labels, as well as rollouts from MPC and the RL policy, are provided in Movie S5.



\section*{DISCUSSION}
\begin{figure*}[t!]
\captionsetup{format=plain}
    \centering
    \makebox[0pt]{\includegraphics[keepaspectratio, width = 
\textwidth]{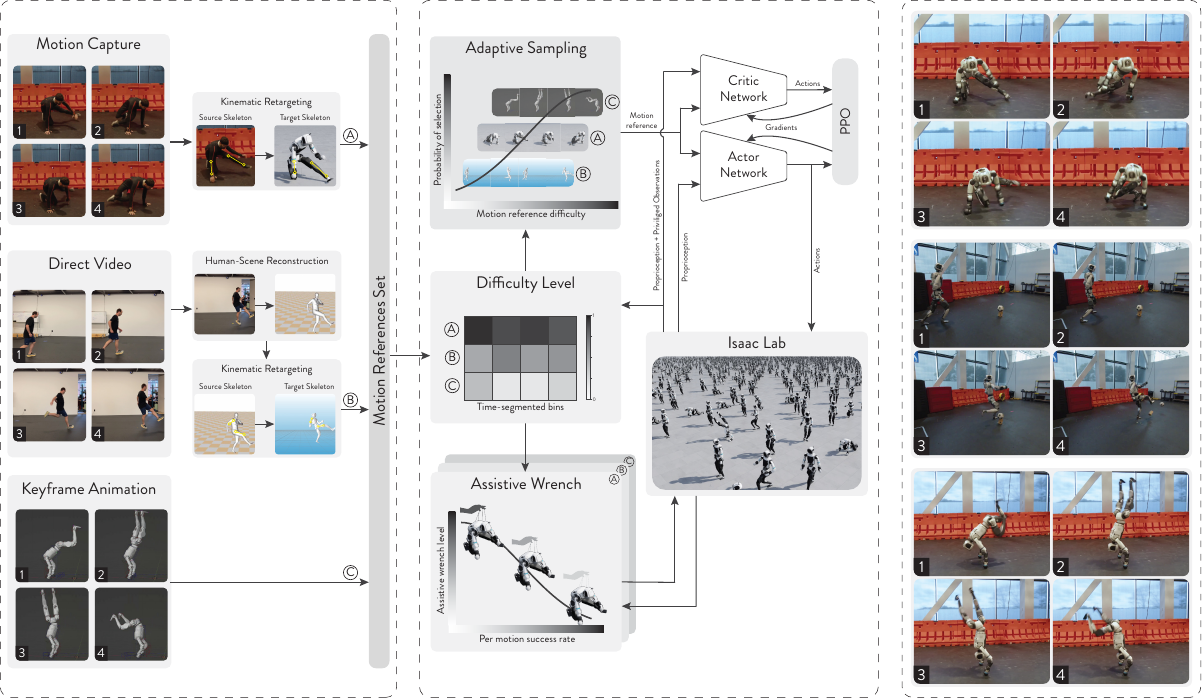}}
    \caption{\textbf{Overview of \emph{ZEST}, which consists of three main stages.} \textbf{(1) Reference data:} A diverse set of motions from MoCap, ViCap, and keyframe animation is processed; MoCap/ViCap references are kinematically retargeted to the target robot. \textbf{(2) Training setup and MDP formulation:} In simulation, the policy is trained using only on-robot signals and the next target state from the reference, while a separate critic receives privileged information (e.g., true base velocity, contact forces) to accelerate learning. To handle long-horizon clips and scale beyond single skills, an adaptive sampling scheme is proposed: trajectories are segmented into fixed-duration bins; a per-bin difficulty level metric is updated via an EMA of failure scores; and a categorical sampler biases reset-state selection toward harder bins while avoiding catastrophic forgetting of easier behaviors. A model-based virtual assistive wrench is applied at the base to stabilize training for highly dynamic behaviors; the current bin's difficulty level modulates its magnitude and is automatically annealed to zero as tracking improves. \textbf{(3) Zero-shot deployment:} The trained policy is deployed directly to the physical robot without any fine-tuning.}    
    \label{fig:method_overview}
\end{figure*}

\noindent \emph{ZEST} bridges motion datasets and robust whole-body control on legged robots. In doing so, it delivers three core contributions. First, it establishes a generic and streamlined motion-imitation recipe that trains RL policies in a single stage from heterogeneous reference sources with various imperfections (high-fidelity MoCap, noisy ViCap, and non-physics-constrained animation) and deploys it on hardware zero-shot. \emph{ZEST} avoids complexities such as contact labels, reference or observation windows, state-estimators, extensive reward shaping, and multi-stage training. Second, our work achieves a new state of the art in demonstrated hardware capabilities across multiple embodiments. On Atlas, we demonstrate, to our knowledge, the first dynamic multi-contact behaviors on a full-size humanoid. Concurrently, on the Unitree G1, we present the first physical motion transfer of highly dynamic skills, such as dancing and box climbing, directly from video. Third, while focused on Atlas, the same policy structure and training recipe carry over to distinct morphologies such as Boston Dynamics' Spot and Unitree's G1. Compared to a state-of-the-art whole-body MPC baseline, which prescribes contacts only at predefined end-effectors, the learned controller eliminates the need for contact schedules and motion-specific costs, is more robust to modeling and contact-timing mismatches, and accommodates interactions with arbitrary body parts (e.g., knees, torso, forearms). 

On the other hand, several limitations remain and motivate our next steps. First, we do not evaluate the generalization of our multi-skill policy to motions outside the training set -- unseen skills that are close to, but not identical with, the training distribution. Such tests would probe whether the policy is able to learn transferable control primitives as opposed to purely overfitting to specific references. We defer this generalization study to future work. Second, the current formulation is proprioceptive and assumes flat and non-slippery terrain; explicit perception of uneven or compliant environments is left for future work. Finally, sim-to-real hinges on reasonable modeling; while we provide a practical procedure involving PLA modeling and a principled selection of armature-dependent PD gains, fully automated system identification remains an open problem. Particularly challenging are complex phenomena that are difficult to capture with first-principles models, and are likely to necessitate data-driven methodologies.

Furthermore, we outline several directions that extend the scope of this work. First, we will pursue general tracking of previously unseen motions via compact motion embeddings or with zero-/few-shot adaptations, paired with continual-learning strategies to expand libraries without catastrophic forgetting. In parallel, we plan to move beyond full-reference control toward partially conditioned interfaces. One approach is to train a teacher to track a large corpus with full references and then distill the knowledge to a student that accepts sparse, human-interpretable inputs. These inputs could include keyframes, short pose snippets, object and scene keyframes, or language commands, following the masked-conditioning paradigm explored in the literature on physics-based character control \cite{Tessler2024MaskedMimic}. Additionally, we will explore high-level generative planners that map user commands to motion plans, which a general-purpose RL-based tracker can execute, thereby closing the loop between behavior synthesis and control, as demonstrated in recent character-control systems \cite{Tevet2025CLoSD,Xu2025PARC}.


\section*{MATERIALS AND METHODS}

\noindent We now present the technical details of \emph{ZEST}. This section details its architecture, as illustrated in Figure~\ref{fig:method_overview}, with a primary focus on the goal-conditioned MDP formulation. We describe the core components of this framework, including modeling, curation of the reference motion dataset, the training setup, and the policy deployment process on hardware. 

\subsection*{Systems and Modeling}
\label{subsec:modeling}
\noindent We evaluated \emph{ZEST} using three robots with different morphologies: the full-scale Boston Dynamics Atlas humanoid ($30$ DoF, $1.8 \mathrm{m}$, $100 \mathrm{kg}$), the smaller Unitree G1 humanoid ($29$ DoF, $1.2 \mathrm{m}$, $35 \mathrm{kg}$), and the Boston Dynamics Spot quadruped ($12$ DoF, $33 \mathrm{kg}$). Our RL approach uses the Isaac Lab simulator \cite{mittal2025isaac}. However, a primary challenge for humanoids is to efficiently model the Parallel-Linkage Actuator (PLA), which is used in ankles, knees, and waists. While a crucial design for dynamic performance, simulating their closed-chain mechanisms results in stiff dynamics that are computationally expensive to calculate. To resolve the trade-off between simulation fidelity and computational efficiency, we developed a series of progressive approximations for the PLAs. A brief overview of these models is provided below, with the approach illustrated in Figure~\ref{fig:closure_approx}. Detailed mathematical derivations are available in the Supplementary Materials (Section S2).

\begin{enumerate}
    \item \textit{Locally Projected Model (Massless-Links Approx.)}: Our first approximation is motivated by the lightweight design of the PLA's support links. We assume these links are massless while retaining the inertia of the motor armatures and the main kinematic chain of PLA. This yields a simulator-compatible model where the effective motor armature is configuration-dependent.
    \item \textit{Dynamic Armature Model (Diagonal Approx.)}: To handle coupled PLAs (e.g., a humanoid's ankles), which produce non-diagonal armature matrices that standard simulators cannot handle, we use a Jacobi approximation. This method approximates the off-diagonal dynamic effects, introducing only a minor, transient error.
    \item \textit{Nominal Armature Model (Fixed-Configuration Approx.)}: Finally, to eliminate the computational overhead of updating armatures at every timestep ($\approx$20\% slowdown), we compute the armature values once at a single, nominal configuration and fix them. This provides a computationally cheap model and a principled basis for designing the robot's PD controller gains.
\end{enumerate}

While the Spot model proposed in \cite{miller2025high} was sufficient for generating most behaviors, it proved insufficient for the more dynamic triple backflip behavior. To bridge the sim-to-real gap, we developed a more detailed actuator model that incorporates the robot's power-limiting algorithm and simulates nonlinear effects, such as motor magnet saturation and torque losses resulting from transmission inefficiencies and friction. Static parameters of the model (e.g., rotor inertia, friction) were obtained from manufacturer specifications. The positive and negative work efficiencies were identified by optimizing the model against real-world data logged from the robot and randomly selected during training within ranges. See Supplementary Materials (Section S3) for derivations.

\begin{figure} [t!]
    \centering
    \includegraphics[width=0.9\linewidth]{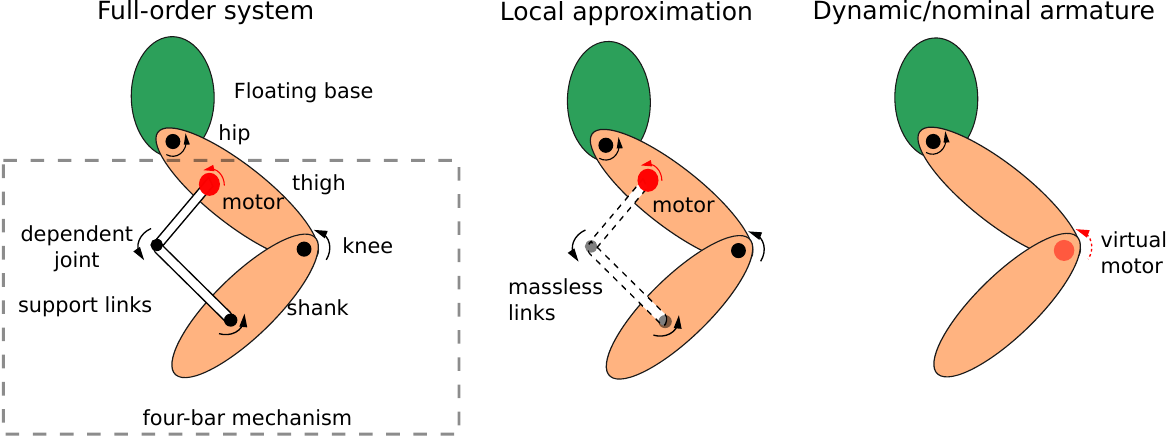}
    \caption{\textbf{Progressive simplification of proposed Parallel-Linkage Actuator models.} It illustrates our modeling approach using a representative humanoid leg, where a motor on the thigh actuates the knee via a four-bar linkage. We begin with \textbf{(1) the Exact Model}, which fully resolves the closed-loop dynamics. We then introduce a series of progressively simplified models. \textbf{(2) The Locally Projected Model} assumes massless support links. \textbf{(3) The Dynamic/Nominal Armature Model} uses a Jacobi approximation for coupled joints, and finally, the most efficient model calculates armature values at a single fixed configuration.}
    \label{fig:closure_approx}
\end{figure}

\subsection*{Reference Motion Pipeline}
\noindent Our framework learns from a diverse library of reference motions, derived from MoCap, ViCap, and keyframe animation. All human-centric data sourced through MoCap or ViCap is mapped to the target robot's skeleton through kinematic retargeting. Crucially, we assume no contact labeling for any data source, relying solely on kinematic information. The following subsections provide a brief overview of our data sources and kinematic retargeting pipeline; comprehensive implementation details are deferred to the Supplementary Materials (Section S4).

\subsubsection*{Data Sources}
\noindent MoCap data, captured using Xsens \cite{xsens_mocap} and Vicon \cite{vicon_mocap} systems, serves as our highest-fidelity source for natural human movements, requiring only minimal post-processing. To leverage the vast amount of motion available in casual videos, our ViCap pipeline reconstructs 3D human motion from a single camera. This process combines \emph{MegaSaM}~\cite{li2025megasam} for robust camera motion and scene estimation with \emph{TRAM}~\cite{wang2024tram} for human pose estimation, a pairing that we found significantly reduces artifacts, such as pose jitter and foot-skidding. Finally, we use Keyframe Animation to create motions that are not humanly possible or are intended for non-humanoid morphologies, such as the Spot quadruped.

\subsubsection*{Kinematic Retargeting}
\noindent To transfer MoCap and ViCap data to a target robot, we solve a spacetime optimization \cite{Gle98}. The objective function comprises several weighted costs: tracking the root position and orientation of the source motion, aligning corresponding bones via alignment frames, and regularizing the output by penalizing joint velocities. During this process, we also jointly optimize for a uniform scale and time resampling of the source data to ensure any ballistic motions are physically consistent with real-world gravity. The optimization is subject to the robot's forward kinematics and non-penetration constraints with itself and the ground.

\subsection*{Training Setup and MDP Formulation}
\noindent We formulate \emph{ZEST} as a goal-conditioned MDP, $\mathcal{M}=(\mathcal{S},\mathcal{A},\mathcal{G},P,r,\gamma)$. The state $s_t \in \mathcal{S}$ represents the robot and environment states, and $a_t \in \mathcal{A}$ is the continuous action corresponding to motor action. At time $t$, the agent receives the observation, $o_t=\psi(s_t,g_t)$ and the target, $g_t \in \mathcal{G}$, selects actions $a_t\!\sim\!\pi_\theta(\,\cdot\,|\,o_t,g_t)$, obtains a reward $r_t=r(s_t,a_t,g_t)$, and the system evolves to a new state $s_{t+1}\!\sim\!P(\,\cdot\,|\,s_t,a_t)$. The goal variable $g_t$ encodes the reference target to be tracked and can be instantiated in several forms: (i) a time-indexed reference snippet $\hat{s}_{t:t+H}$ with $H\!\ge\!0$ (e.g., a single step or a short window); (ii) a latent embedding $z_t=f(\hat{s}_{t:t+H})$ of such a snippet; or (iii) a phase variable $\phi_t\in[0,1]$ plus a trajectory identifier in the case of multiple reference motions. The policy parameters $\theta$ are optimized to maximize the expected discounted return over a horizon $T$
\[
J(\theta)\;=\;\mathbb{E}\!\left[\sum_{t=0}^{T-1}\gamma^{t}\, r(s_t,a_t,g_t)\right],
\]
where the expectation is over rollouts induced by $P$ and $\pi_\theta$, randomized initial states, and the goal distribution.

Training is conducted in Isaac Lab using Proximal Policy Optimization (PPO) \cite{PPO} as the RL algorithm. To stabilize optimization and improve convergence, all observations are normalized using an empirical running-average scheme. The actor and critic networks are modeled as Multi-Layer Perceptrons (MLPs) with three hidden layers and ELU activations.
To accelerate training while preserving the hardware deployability, we utilize an asymmetric actor-critic architecture \cite{AsymmetricActorCritic}. The critic is trained using extra, privileged information available only in simulations, which enables it to estimate the value function more accurately. In contrast, the policy is limited to the observations available during hardware deployment, ensuring it can be transferred without any state estimator. This approach prevents lag and bias from estimator fusion, sidesteps the fragile contact and terrain assumptions often required by state estimators, and eliminates the need for added complexity that is not captured during training. With the learning algorithm established, we now define the specifics of the MDP.

\subsubsection*{Actor and Critic Observations}
\noindent The Actor, otherwise known as policy, receives proprioception information along with target references 
\[
\mathbf{o}_t=(\mathbf{o}_{\text{prop}},\ \mathbf{o}_{\text{ref}}),
\]
the proprioceptive part ${\mathbf{o}_{\text{prop}}=\big({}_T \bm \omega_{IT},\ {}_T \bm g_{I},\ \bm q_j,\ \dot{\bm q}_j,\ \bm a_{t-1}\big)}$
 contains only instantaneous onboard measurements. We intentionally exclude quantities that require a state estimator, such as global base pose or linear velocity. Here, ${}_T \bm \omega_{IT}$ is the absolute angular velocity of the torso-attached IMU frame $\{\mathcal{T}\}$, ${}_T \bm g_{I}$ is the normalized gravity vector expressed in $\{\mathcal{T}\}$, and $(\bm q_j,\dot{\bm q}_j,\bm a_{t-1})$ denote joint positions, joint velocities, and the previous action. We hypothesize that including $\bm a_{t-1}$ is sufficient to infer essential information, such as contact states and base linear velocity, under nominal terrain conditions without resorting to long observation histories; it also provides the minimal temporal context needed to enforce action smoothness. 

The reference observations ${\mathbf{o}_{\text{ref}}=\big({}_I \hat{\bm r}^z_{IB},\ {}_B \hat{\bm v}_{IB},\ {}_B \hat{\bm \omega}_{IB},\ {}_B \hat{\bm g}_{I},\ \hat{\bm q}_j\big)}$ encode the next target state extracted from the demonstration: ${}_I \hat{\bm r}^z_{IB}$ is the base height; ${}_B \hat{\bm v}_{IB}$ and ${}_B \hat{\bm \omega}_{IB}$ are the absolute linear and angular base velocities expressed in the base frame $\{\mathcal{B}\}$; ${}_B \hat{\bm g}_{I}$ is the gravity direction in $\{\mathcal{B}\}$; and $\hat{\bm q}_j$ are the reference joint positions. We deliberately avoid future-reference windows beyond this immediate next target as our ablations showed no benefit for performance or convergence speed. 

The Critic receives the policy observations in addition to extra signals referred to as privileged information. This privileged input includes the base linear velocity, base height, end-effector positions, velocities, and contact forces, as well as curriculum-related signals. Additional details are provided in the Supplementary  (Table S3). 

\subsubsection*{Actions and Joint-Level Control}
\noindent The policy outputs residual actions $\bm a_t$, which are added to the reference joint positions before sending them as position targets to the joint-level PD controllers:
\[
\bm{q}^{\mathrm{cmd}}_j = \hat{\bm{q}}_j + \bm{\Sigma}\,\bm{a}_t,
\]
where $\bm{\Sigma}$ is a positive-definite diagonal matrix of per-DoF action scales. These scales set the amplitude of residual corrections and exploration on each axis and reflect whether additional feedforward torque is needed for reference tracking: actuators that require more assistance (e.g., hips and knees) receive larger scales, whereas those adequately handled by the PD control loop alone (e.g., the robot’s head and wrists) use smaller scales. The PD gains are chosen by modeling each joint as an independent second-order system:
\[
I\,\theta''(t) + K_d\,\theta'(t) + K_p\,\theta(t) = 0,
\]
where $I$ is the nominal armature value (refer to the Modeling section) about that axis. The gains \(K_p\) and \(K_d\) are heuristically tuned to achieve a critically damped response with a desired natural frequency $\omega_n>0$,
\[
K_p = I\,\omega_n^2, \qquad K_d = 2\,I\,\omega_n.
\]

In practice, we set $\omega_n$ to balance between responsiveness and robustness. It is set high enough for fast tracking, but not so high that the loop becomes brittle during deployment or that simulation requires very small time steps to resolve stiff dynamics with fast transients, which would in turn increase training times. The same PD gains are used in simulation and on hardware to minimize sim-to-real mismatch.

\subsubsection*{Reward Terms}
\noindent The total reward combines three main components: a tracking reward, a regularization reward, and a survival reward,
\begin{equation}
    r_{\text{total}} \;=\; r_{\text{track}} + r_{\text{reg}} + r_{\text{survival}}.
\end{equation}
All terms are deliberately kept generic: nothing is task- or motion-specific, and the reward setup does not rely on reference contact labels from the demonstrations, making it broadly applicable across behaviors. The tracking reward measures how well the policy follows the reference motion. It can be expressed in different forms, such as a quadratic penalty or a decaying exponential kernel; in our case, we adopt the latter, which yields a dense, bounded shaping signal that emphasizes accuracy near the reference and varies smoothly with the tracking error:
\begin{equation}
    r_{\text{track}} 
    \;=\; 
    \sum_{i} c_{t_i} \exp\!\Bigl(-\,\kappa\frac{\|\mathbf{e}_i\|^2}{\sigma_i^2}\Bigr),
\end{equation}
where $c_{t_i}$ is the reward weight for term $i$, and $\mathbf{e}_i$ denotes the error between the current and reference values for quantities such as base linear velocity, base angular velocity, base pose, joint positions, and keybody poses. The regularization component, denoted $r_{\text{reg}}$, aggregates penalties that encourage smooth and physically feasible behavior. Concretely, we penalize rapid changes in actions across time steps, large joint accelerations, and violations of joint position and torque limits. The survival reward is a constant positive term granted at each step, $r_{\text{survival}} = c_{\text{survival}}$, which discourages premature termination by rewarding longer episodes. Full functional forms and hyperparameters for all reward terms are provided in the Supplementary Materials (Table S4).


\subsubsection*{Early Terminations}
\noindent To steer exploration toward promising states and avoid wasting samples on highly undesirable trajectories, we trigger early termination under conditions such as contact forces exceeding a predefined threshold or a large deviation of the agent from the reference.


\subsubsection*{Domain Randomization}
\noindent To ensure robust zero-shot sim-to-real transfer, we employ domain randomization to prevent the policy from overfitting to the nominal simulation model. Our domain randomization strategy includes injecting Gaussian noise into observations, applying random impulsive pushes to the robot at random times, varying surface friction coefficient, and slightly randomizing link masses around their nominal values. We used these randomizations carefully, as excessive variability can degrade performance by making policies overly conservative or aggressive. The details of the domain randomization's hyperparameters can be found in the Supplementary  (Table S5).

\subsubsection*{Adaptive Reference State Initialization}
\noindent To initialize our environment during training, we first adopt the Reference State Initialization (RSI) strategy introduced in \cite{Peng2018DeepMimic}. At each reset, a random phase $\phi_{\text{init}}\!\sim\!\mathcal{U}(0,1)$ is sampled. Then the environment is initialized at the corresponding reference state $\hat{s}(\phi_{\text{init}})$. RSI exposes the policy to relevant states along the trajectory from the outset, decouples episode length from demonstration length, and improves sample efficiency. Additionally, when a rollout reaches the end of a demonstration, we insert a brief dwell and then reset by timeout (not failure), which further increases the number of informative transitions. 

While uniform RSI is generally effective, with long-horizon trajectories or large sets of trajectories, it can be inefficient because it blindly samples phases, oversampling already-mastered regions and undersampling difficult segments. To bias sampling toward where learning is most needed, we use an adaptive RSI scheme: all demonstrations are partitioned into fixed $N$-step bins, and each bin $b$ maintains a failure level $f_b$ derived from recent tracking performance (based on a similarity score metric) and updated with an EMA filter. At reset, a bin index $k$ is drawn from a categorical distribution whose logits are proportional to $\{f_b\}$, and a phase is then sampled uniformly within the selected bin to obtain $\phi_{\text{init}}$; a small floor probability is retained for every bin to prevent forgetting previously acquired skills. This procedure increases sampling density on difficult motions and difficult phases while preserving global coverage; additional implementation details are provided in the Supplementary  (Section S5).


\subsubsection*{Assistive Wrench Automatic Curriculum}
\noindent While the framework presented so far is sufficient to track most motions in our library, skills with highly dynamic base rotations and large angular rates (e.g., handstands, cartwheels, backflips) tend to trigger immediate early terminations under RSI alone and converge significantly slower compared to simpler behaviors such as walking and running. To address this, we employ an automatic assistive-wrench curriculum, serving as a continuation strategy that gradually increases task difficulty: a virtual spatial wrench is applied at the base and annealed as tracking improves, analogous to a gymnastics coach providing diminishing support to their student. Similar assistive-force curricula have been explored in previous related works \cite{Karpathy2012Curriculum, Yu2018Symmetric, HoST2025, Cao2025A2CF}. The assistive wrench is computed in a model-based fashion using a PD term on the base pose tracking error, together with a feedforward component that compensates nominal torso dynamics. To keep assistance partially supportive while still compelling the agent to learn the skill for itself, we scale the assistive wrench by a gain ${\beta\!\in\!(0,\beta_{\text{max}})}$ with ${\beta_{\text{max}} < 1.0}$. The annealing schedule is driven automatically by the per-bin failure levels introduced in the adaptive RSI procedure: bins with higher failure receive stronger assistance initially and are relaxed more aggressively as tracking improves. Assistance decays quickly and ultimately vanishes once a target tracking performance is reached. Additional details are provided in the Supplementary  (Section S6).



\subsection*{Hardware Deployment Pipeline}
\noindent An important yet often overlooked aspect of embodied intelligence is the hardware deployment pipeline. Since our research involves frequent testing on physical robots, we created an efficient, automated, framework- and hardware-agnostic pipeline to ensure safe and consistent releases across multiple platforms. This pipeline follows a three-stage workflow. First, during training, policy checkpoints are exported in ONNX format and logged to a registry, along with detailed metadata that captures all relevant configurations. For validation, these artifacts are retrieved, and the metadata is used to automatically configure the controller in our \emph{evaluation simulator} \cite{todorov2012mujoco}, which mirrors the hardware's software stack. Once a policy performs as expected, it is deployed to the physical robot using the exact same automated process, guaranteeing consistency and eliminating manual setup errors. It is worth noting that while some deployment workflow is necessary for hardware testing, the learned policy is not tied to this specific pipeline; we include it primarily to accelerate iteration and reduce incidental sources of error during sim-to-sim evaluation and hardware testing.

\section*{Acknowledgments}
We would like to thank Merritt Moore for the MoCap data collection. Moreover, we note that while all ideas, methods, and results are original to the authors, AI-based tools were used solely to refine the clarity and readability of the manuscript text. \textbf{Author contributions:} The \textit{Core Contributors} designed and implemented the ZEST pipeline; trained most of the policies, and oversaw the hardware experiments on robots. They also contributed to the writing of the manuscript. The \textit{Additional Contributors} assisted with engineering tasks to support the project, including deployment code on hardware, data collection and curation, as well as supporting hardware experiments. The \textit{Project Leads} set the overarching vision by defining the core scientific goals. They provided technical guidance and coordinated the team’s efforts. Finally, they helped shape the overall narrative of the paper. \textbf{Competing interests:} The authors declare that they have no competing interests. \textbf{Data and materials availability:} All data needed to evaluate the conclusions in the paper are present in the paper or the Supplementary Materials.

\section*{Supplementary materials}
\vspace{1ex}

{
\renewcommand{\arraystretch}{0.9}
\noindent
\begin{tabular}{@{}p{1.4cm}p{\dimexpr\linewidth-1.4cm\relax}@{}}
Section S1. & Nomenclature \\
Section S2. & Implementation Details: Actuator Modeling \\
Section S3. & Implementation Details: Spot Modeling \\
Section S4. & Implementation Details: Motion Dataset \\
Section S5. & Implementation Details: Adaptive RSI Sampling \\
Section S6. & Implementation Details: Assistive Wrench Curriculum \\
Figure S1. & Illustration of Main Chain and Support Chain \\
Figure S2. & Atlas Ankle Torque Limits Profile \\
Figure S3. & Effect of Reference and Observation Windows on Success.\\
Figure S4. & Robustness to Model Uncertainty and Disturbances. \\
Figure S5. & Robustness Maps from Varied Initial Conditions. \\
Table S1. & Library of Reference Motions for Multi-Skill Policy \\
Table S2. & Normalized MSE of Simplified PLA Models. \\
Table S3. & Observation Terms Summary \\
Table S4. & Reward Terms Summary \\
Table S5. & Domain Randomization Summary \\
Table S6. & MDP Hyperparameters \\
Table S7. & PPO Hyperparameters \\
\end{tabular}
}

\vspace{1.5ex}

\noindent\textbf{Other Supplementary Material for this manuscript includes:}

\vspace{0.5ex}

{
\renewcommand{\arraystretch}{0.9}
\noindent
\begin{tabular}{@{}p{1.4cm}p{\dimexpr\linewidth-1.4cm\relax}@{}}
\href{https://youtu.be/y9xzSBZVGGA}{Movie S1.} & Retargeted MoCap-derived References \\
\href{https://youtu.be/8JBW1Mowx_8}{Movie S2.} & Retargeted ViCap-derived References \\
\href{https://youtu.be/BT9QzAKoBUw}{Movie S3.} & Keyframe Animation References \\
\href{https://youtu.be/8TNO7Mzjg9k}{Movie S4.} & Simulation-based Multi-Skill Policy Evaluation \\
\href{https://youtu.be/sCse8_MSN1w}{Movie S5.} & Simulation-based Benchmarking against MPC \\
\href{https://youtu.be/fmWCCcRuHOw}{Movie S6.} & Box-Climbing: Repeatability Testing on Hardware.  \\
\href{https://youtu.be/OD0K6EYF7f8}{Movie S7.} & Box-Climbing: Robustness Testing on Hardware. \\
\href{https://youtu.be/Wvw-ss0qIuI}{Movie S8.} & Impact of PLA Modeling on Sim-to-Real Transfer. \\
\end{tabular}
}

\bibliography{scibib}

\begin{thebibliography}{69}
\providecommand{\natexlab}[1]{#1}
\providecommand{\url}[1]{\texttt{#1}}
\expandafter\ifx\csname urlstyle\endcsname\relax
  \providecommand{\doi}[1]{doi: #1}\else
  \providecommand{\doi}{doi: \begingroup \urlstyle{rm}\Url}\fi

\bibitem[Kuindersma et~al.(2016)Kuindersma, Deits, Fallon, Valenzuela, Dai, Permenter, Koolen, Marion, and Tedrake]{Kuindersma2016AtlasWBC}
Scott Kuindersma, Robin Deits, Maurice~F. Fallon, Andres Valenzuela, Hongkai Dai, Frank Permenter, Twan Koolen, Pat Marion, and Russ Tedrake.
\newblock Optimization-based locomotion planning, estimation, and control design for the atlas humanoid robot.
\newblock \emph{Auton. Robots}, 40\penalty0 (3):\penalty0 429--455, 2016.
\newblock \doi{10.1007/s10514-015-9479-3}.

\bibitem[{Boston Dynamics}(2024{\natexlab{a}})]{BostonDynamics2024PickingUpMomentum}
{Boston Dynamics}.
\newblock Picking up momentum.
\newblock Boston Dynamics Blog, 2024{\natexlab{a}}.
\newblock URL \url{https://bostondynamics.com/blog/picking-up-momentum/}.
\newblock Blog post on Model Predictive Control for Atlas; accessed 2025-08-12.

\bibitem[Kuindersma(2020)]{Kuindersma2020RecentProgressAtlas}
Scott Kuindersma.
\newblock Recent progress on atlas, the world’s most dynamic humanoid robot.
\newblock YouTube video, 2020.
\newblock URL \url{https://www.youtube.com/watch?v=EGABAx52GKI}.

\bibitem[{Boston Dynamics}(2023)]{AtlasGetsAGrip2025}
{Boston Dynamics}.
\newblock Atlas gets a grip.
\newblock YouTube video, 2023.
\newblock URL \url{https://www.youtube.com/watch?v=-e1_QhJ1EhQ}.

\bibitem[{Boston Dynamics}(2024{\natexlab{b}})]{HDAtlasManipulates}
{Boston Dynamics}.
\newblock Hd atlas manipulates | boston dynamics.
\newblock YouTube video, 2024{\natexlab{b}}.
\newblock URL \url{https://www.youtube.com/watch?v=LeeiN9smjjY}.

\bibitem[Kudruss et~al.(2015)Kudruss, Naveau, Stasse, Mansard, Kirches, Sou{\`e}res, and Mombaur]{KudrussOCP}
Manuel Kudruss, Maximilien Naveau, Olivier Stasse, Nicolas Mansard, Christian Kirches, Philippe Sou{\`e}res, and Katja~D. Mombaur.
\newblock Optimal control for whole-body motion generation using center-of-mass dynamics for predefined multi-contact configurations.
\newblock In \emph{IEEE-RAS Int. Conf. Humanoid Robots (Humanoids)}, pages 684--689, 2015.
\newblock \doi{10.1109/HUMANOIDS.2015.7363428}.

\bibitem[Li and Wensing(2025)]{li2024cafe}
He~Li and Patrick~M. Wensing.
\newblock Cafe-mpc: A cascaded-fidelity model predictive control framework with tuning-free whole-body control.
\newblock \emph{IEEE Trans. Robotics}, 41:\penalty0 837--856, 2025.
\newblock \doi{10.1109/TRO.2024.3504132}.

\bibitem[Grandia et~al.(2023)Grandia, Farshidian, Knoop, Schumacher, Hutter, and B{\"a}cher]{RubenDOC}
Ruben Grandia, Farbod Farshidian, Espen Knoop, Christian Schumacher, Marco Hutter, and Moritz B{\"a}cher.
\newblock {DOC}: Differentiable optimal control for retargeting motions onto legged robots.
\newblock \emph{ACM Trans. Graph.}, 42\penalty0 (4):\penalty0 96:1--96:14, 2023.
\newblock \doi{10.1145/3592454}.

\bibitem[Sleiman et~al.(2023)Sleiman, Farshidian, and Hutter]{Sleiman2023MultiContact}
Jean-Pierre Sleiman, Farbod Farshidian, and Marco Hutter.
\newblock Versatile multi-contact planning and control for legged loco-manipulation.
\newblock \emph{Science Robotics}, 8\penalty0 (81):\penalty0 eadg5014, 2023.
\newblock \doi{10.1126/scirobotics.adg5014}.

\bibitem[Rudin et~al.(2021)Rudin, Hoeller, Reist, and Hutter]{RL_Nikita}
Nikita Rudin, David Hoeller, Philipp Reist, and Marco Hutter.
\newblock Learning to walk in minutes using massively parallel deep reinforcement learning.
\newblock In \emph{Conference on Robot Learning (CoRL)}, volume 164 of \emph{Proceedings of Machine Learning Research}, pages 91--100. PMLR, 2021.

\bibitem[Zhao et~al.(2020)Zhao, Queralta, and Westerlund]{RL_SimToRealSurvey}
Wenshuai Zhao, Jorge~Pe{\~n}a Queralta, and Tomi Westerlund.
\newblock Sim-to-real transfer in deep reinforcement learning for robotics: A survey.
\newblock In \emph{IEEE Symposium Series on Computational Intelligence (SSCI)}, pages 737--744, 2020.
\newblock \doi{10.1109/SSCI47803.2020.9308468}.

\bibitem[Ha et~al.(2025)Ha, Lee, van~de Panne, Xie, Yu, and Khadiv]{RLLocomotionSurvey}
Sehoon Ha, Joonho Lee, Michiel van~de Panne, Zhaoming Xie, Wenhao Yu, and Majid Khadiv.
\newblock Learning-based legged locomotion: State of the art and future perspectives.
\newblock \emph{Int. J. Robot. Res.}, 44\penalty0 (8):\penalty0 1396--1427, 2025.
\newblock \doi{10.1177/02783649241312698}.

\bibitem[Gu et~al.(2025)Gu, Li, Shen, Yu, Xie, McCrory, Cheng, Shamsah, Griffin, Liu, et~al.]{RLHumanoidSurvey}
Zhaoyuan Gu, Junheng Li, Wenlan Shen, Wenhao Yu, Zhaoming Xie, Stephen McCrory, Xianyi Cheng, Abdulaziz Shamsah, Robert Griffin, C~Karen Liu, et~al.
\newblock Humanoid locomotion and manipulation: Current progress and challenges in control, planning, and learning.
\newblock \emph{IEEE/ASME Trans. Mechatronics}, 2025.

\bibitem[Lee et~al.(2020)Lee, Hwangbo, Wellhausen, Koltun, and Hutter]{RL_Joonho}
Joonho Lee, Jemin Hwangbo, Lorenz Wellhausen, Vladlen Koltun, and Marco Hutter.
\newblock Learning quadrupedal locomotion over challenging terrain.
\newblock \emph{Science Robotics}, 5\penalty0 (47):\penalty0 eabc5986, 2020.
\newblock \doi{10.1126/scirobotics.abc5986}.

\bibitem[Miki et~al.(2022)Miki, Lee, Hwangbo, Wellhausen, Koltun, and Hutter]{Miki2022WildLocomotion}
Takahiro Miki, Joonho Lee, Jemin Hwangbo, Lorenz Wellhausen, Vladlen Koltun, and Marco Hutter.
\newblock Learning robust perceptive locomotion for quadrupedal robots in the wild.
\newblock \emph{Science Robotics}, 7\penalty0 (62):\penalty0 eabk2822, 2022.
\newblock \doi{10.1126/scirobotics.abk2822}.

\bibitem[Sun et~al.(2025)Sun, Cao, Chen, Su, Liu, Xie, and Liu]{HumanoidPerceptiveLocomotion}
Wandong Sun, Baoshi Cao, Long Chen, Yongbo Su, Yang Liu, Zongwu Xie, and Hong Liu.
\newblock Learning perceptive humanoid locomotion over challenging terrain.
\newblock \emph{arXiv preprint arXiv:2503.00692}, 2025.

\bibitem[Miller et~al.(2025{\natexlab{a}})Miller, Yu, Brauckmann, and Farshidian]{AJSpot}
A.J. Miller, Fangzhou Yu, Michael Brauckmann, and Farbod Farshidian.
\newblock High-performance reinforcement learning on spot: Optimizing simulation parameters with distributional measures.
\newblock In \emph{IEEE Int. Conf. on Robotics and Automation (ICRA)}, pages 9981--9988, 2025{\natexlab{a}}.
\newblock \doi{10.1109/ICRA55743.2025.11128575}.

\bibitem[Bellegarda et~al.(2022)Bellegarda, Chen, Liu, and Nguyen]{HighSpeedRunning}
Guillaume Bellegarda, Yiyu Chen, Zhuochen Liu, and Quan Nguyen.
\newblock Robust high-speed running for quadruped robots via deep reinforcement learning.
\newblock In \emph{2022 IEEE/RSJ International Conference on Intelligent Robots and Systems (IROS)}, pages 10364--10370, 2022.
\newblock \doi{10.1109/IROS47612.2022.9982132}.

\bibitem[Cheng et~al.(2024)Cheng, Shi, Agarwal, and Pathak]{Cheng2023ExtremeParkour}
Xuxin Cheng, Kexin Shi, Ananye Agarwal, and Deepak Pathak.
\newblock Extreme parkour with legged robots.
\newblock In \emph{2024 IEEE International Conference on Robotics and Automation (ICRA)}, pages 11443--11450, 2024.
\newblock \doi{10.1109/ICRA57147.2024.10610200}.

\bibitem[Hoeller et~al.(2024)Hoeller, Rudin, Sako, and Hutter]{Hoeller2024AnymalParkour}
David Hoeller, Nikita Rudin, Dhionis~V Sako, and Marco Hutter.
\newblock Anymal parkour: Learning agile navigation for quadrupedal robots.
\newblock \emph{Science Robotics}, 9\penalty0 (88):\penalty0 eadi7566, 2024.
\newblock \doi{10.1126/scirobotics.adi7566}.

\bibitem[Rudin et~al.(2025)Rudin, He, Aurand, and Hutter]{Rudin2025ParkourWild}
Nikita Rudin, Junzhe He, Joshua Aurand, and Marco Hutter.
\newblock Parkour in the wild: Learning a general and extensible agile locomotion policy using multi-expert distillation and rl fine-tuning.
\newblock \emph{arXiv preprint arXiv:2505.11164}, 2025.

\bibitem[{OpenAI} et~al.(2020){OpenAI}, Andrychowicz, Baker, Chociej, Józefowicz, McGrew, Pachocki, Petron, Plappert, Powell, Ray, Schneider, Sidor, Hessel, Agarwal, Henaff, Pinto, Abbeel, and Zaremba]{Andrychowicz2020DexterousManipulation}
{OpenAI}, Marcin Andrychowicz, Bowen Baker, Maciek Chociej, Rafał Józefowicz, Bob McGrew, Jakub Pachocki, Arthur Petron, Matthias Plappert, Glenn Powell, Alex Ray, Jonas Schneider, Szymon Sidor, Jack Hessel, Rishabh Agarwal, Mikael Henaff, Lerrel Pinto, Pieter Abbeel, and Wojciech Zaremba.
\newblock Learning dexterous in-hand manipulation.
\newblock \emph{The International Journal of Robotics Research}, 39\penalty0 (1):\penalty0 3--20, 2020.
\newblock \doi{10.1177/0278364919887447}.

\bibitem[Handa et~al.(2023)Handa, Allshire, Makoviychuk, Petrenko, Singh, Liu, Makoviichuk, Van~Wyk, Zhurkevich, Sundaralingam, Narang, Lafleche, Fox, and State]{Handa2023Dextreme}
Ankur Handa, Arthur Allshire, Viktor Makoviychuk, Aleksei Petrenko, Ritvik Singh, Jingzhou Liu, Denys Makoviichuk, Karl Van~Wyk, Alexander Zhurkevich, Balakumar Sundaralingam, Yashraj Narang, Jean-Francois Lafleche, Dieter Fox, and Gavriel State.
\newblock {DeXtreme}: Transfer of agile in-hand manipulation from simulation to reality.
\newblock In \emph{2023 IEEE International Conference on Robotics and Automation (ICRA)}, pages 5962--5969, 2023.
\newblock \doi{10.1109/ICRA48891.2023.10160216}.

\bibitem[Lin et~al.(2025)Lin, Sachdev, Fan, Malik, and Zhu]{HumanoidDexterousManipulation}
Toru Lin, Kartik Sachdev, Linxi Fan, Jitendra Malik, and Yuke Zhu.
\newblock Sim-to-real reinforcement learning for vision-based dexterous manipulation on humanoids.
\newblock \emph{arXiv preprint arXiv:2502.20396}, 2025.

\bibitem[Ma et~al.(2025)Ma, Cramariuc, Farshidian, and Hutter]{Ma2025Badminton}
Yuntao Ma, Andrei Cramariuc, Farbod Farshidian, and Marco Hutter.
\newblock Learning coordinated badminton skills for legged manipulators.
\newblock \emph{Science Robotics}, 10\penalty0 (102):\penalty0 eadu3922, 2025.
\newblock \doi{10.1126/scirobotics.adu3922}.

\bibitem[Zhang et~al.(2025{\natexlab{a}})Zhang, Yuan, Gurunath, He, Omidshafiei, Agha-mohammadi, Vazquez-Chanlatte, Pedersen, and Shi]{FALCON}
Yuanhang Zhang, Yifu Yuan, Prajwal Gurunath, Tairan He, Shayegan Omidshafiei, Ali-akbar Agha-mohammadi, Marcell Vazquez-Chanlatte, Liam Pedersen, and Guanya Shi.
\newblock Falcon: Learning force-adaptive humanoid loco-manipulation.
\newblock \emph{arXiv preprint arXiv:2505.06776}, 2025{\natexlab{a}}.

\bibitem[Dadiotis et~al.(2025)Dadiotis, Mittal, Tsagarakis, and Hutter]{DynamicBoxPushing}
Ioannis Dadiotis, Mayank Mittal, Nikos Tsagarakis, and Marco Hutter.
\newblock Dynamic object goal pushing with mobile manipulators through model-free constrained reinforcement learning.
\newblock In \emph{2025 IEEE International Conference on Robotics and Automation (ICRA)}, pages 13363--13369, 2025.
\newblock \doi{10.1109/ICRA55743.2025.11128166}.

\bibitem[Schwarke et~al.(2023)Schwarke, Klemm, van~der Boon, Bjelonic, and Hutter]{CuriosityRL}
Clemens Schwarke, Victor Klemm, Matthijs van~der Boon, Marko Bjelonic, and Marco Hutter.
\newblock Curiosity-driven learning of joint locomotion and manipulation tasks.
\newblock In \emph{Conference on Robot Learning (CoRL)}, volume 229 of \emph{Proceedings of Machine Learning Research}, pages 2594--2610. PMLR, 2023.

\bibitem[Peng et~al.(2018)Peng, Abbeel, Levine, and van~de Panne]{Peng2018DeepMimic}
Xue~Bin Peng, Pieter Abbeel, Sergey Levine, and Michiel van~de Panne.
\newblock {DeepMimic}: Example-guided deep reinforcement learning of physics-based character skills.
\newblock \emph{ACM Trans. Graph.}, 37\penalty0 (4):\penalty0 143:1--143:14, 2018.
\newblock \doi{10.1145/3197517.3201311}.

\bibitem[Bergamin et~al.(2019)Bergamin, Clavet, Holden, and Forbes]{Bergamin2019DReCon}
Kevin Bergamin, Simon Clavet, Daniel Holden, and James~Richard Forbes.
\newblock {DReCon}: Data-driven responsive control of physics-based characters.
\newblock \emph{ACM Trans. Graph.}, 38\penalty0 (6):\penalty0 206:1--206:11, 2019.
\newblock \doi{10.1145/3355089.3356536}.

\bibitem[Yuan and Kitani(2020)]{Yuan2020RFC}
Ye~Yuan and Kris~M. Kitani.
\newblock Residual force control for agile human behavior imitation and extended motion synthesis.
\newblock In Hugo Larochelle, Marc'Aurelio Ranzato, Raia Hadsell, Maria-Florina Balcan, and Hsuan-Tien Lin, editors, \emph{Advances in Neural Information Processing Systems 33}, pages 21763--21774, 2020.
\newblock \doi{10.5555/3495724.3497550}.

\bibitem[Luo et~al.(2023)Luo, Cao, Winkler, Kitani, and Xu]{Luo2023PHC}
Zhengyi Luo, Jinkun Cao, Alexander~W. Winkler, Kris~M. Kitani, and Weipeng Xu.
\newblock Perpetual humanoid control for real-time simulated avatars.
\newblock In \emph{Proceedings of the IEEE/CVF International Conference on Computer Vision (ICCV)}, pages 10895--10904, 2023.
\newblock \doi{10.1109/ICCV51070.2023.01000}.

\bibitem[Peng et~al.(2021)Peng, Ma, Abbeel, Levine, and Kanazawa]{Peng2021AMP}
Xue~Bin Peng, Ze~Ma, Pieter Abbeel, Sergey Levine, and Angjoo Kanazawa.
\newblock {AMP}: Adversarial motion priors for stylized physics-based character control.
\newblock \emph{ACM Trans. Graph.}, 40\penalty0 (4):\penalty0 144:1--144:20, 2021.
\newblock \doi{10.1145/3450626.3459670}.

\bibitem[Peng et~al.(2022)Peng, Guo, Halper, Levine, and Fidler]{Peng2022ASE}
Xue~Bin Peng, Yunrong Guo, Lina Halper, Sergey Levine, and Sanja Fidler.
\newblock {ASE}: Large-scale reusable adversarial skill embeddings for physically simulated characters.
\newblock \emph{ACM Trans. Graph.}, 41\penalty0 (4):\penalty0 94:1--94:17, 2022.
\newblock \doi{10.1145/3528223.3530110}.

\bibitem[Tessler et~al.(2023)Tessler, Kasten, Guo, Mannor, Chechik, and Peng]{Tessler2023CALM}
Chen Tessler, Yoni Kasten, Yunrong Guo, Shie Mannor, Gal Chechik, and Xue~Bin Peng.
\newblock {CALM}: Conditional adversarial latent models for directable virtual characters.
\newblock In \emph{ACM SIGGRAPH 2023 Conference Proceedings}, 2023.
\newblock \doi{10.1145/3588432.3591541}.

\bibitem[Tevet et~al.(2025)Tevet, Kasten, Tessler, Gabel, Bermano, and Cohen-Or]{Tevet2025CLoSD}
Guy Tevet, Yoni Kasten, Chen Tessler, Moshe Gabel, Amit~H. Bermano, and Daniel Cohen-Or.
\newblock {CLoSD}: Closing the loop between simulation and diffusion for multi-task character control.
\newblock In \emph{International Conference on Learning Representations (ICLR)}, 2025.

\bibitem[Xu et~al.(2025{\natexlab{a}})Xu, Shi, Yin, and Peng]{Xu2025PARC}
Michael Xu, Yi~Shi, KangKang Yin, and Xue~Bin Peng.
\newblock Parc: Physics-based augmentation with reinforcement learning for character controllers.
\newblock In \emph{SIGGRAPH 2025 Conference Papers}, 2025{\natexlab{a}}.
\newblock \doi{10.1145/3721238.3730616}.

\bibitem[Xu et~al.(2025{\natexlab{b}})Xu, Ling, Wang, and Gui]{Xu2025InterMimic}
Zhengyi Xu, Hung~Yu Ling, Yu-Xiong Wang, and Liang-Yan Gui.
\newblock Intermimic: Towards universal whole-body control for physics-based human-object interactions.
\newblock In \emph{IEEE/CVF Conf. on Computer Vision and Pattern Recognition (CVPR)}, 2025{\natexlab{b}}.

\bibitem[Peng et~al.(2020)Peng, Coumans, Zhang, Lee, Tan, and Levine]{Peng2020ImitateAnimals}
Xue~Bin Peng, Erwin Coumans, Tingnan Zhang, Tsang-Wei~Edward Lee, Jie Tan, and Sergey Levine.
\newblock Learning agile robotic locomotion skills by imitating animals.
\newblock In \emph{Robotics: Science and Systems (RSS)}, 2020.

\bibitem[Grandia et~al.(2024)Grandia, Knoop, Hopkins, Wiedebach, Bishop, Pickles, M{\"u}ller, and B{\"a}cher]{Grandia2024BipedalCharacter}
Ruben Grandia, Espen Knoop, Michael~A. Hopkins, Georg Wiedebach, Jared Bishop, Steven Pickles, David M{\"u}ller, and Moritz B{\"a}cher.
\newblock Design and control of a bipedal robotic character.
\newblock In \emph{Robotics: Science and Systems (RSS)}, 2024.

\bibitem[Allshire et~al.(2025)Allshire, Choi, Zhang, McAllister, Zhang, Kim, Darrell, Abbeel, Malik, and Kanazawa]{Allshire2025VideoMimic}
Arthur Allshire, Hongsuk Choi, Junyi Zhang, David McAllister, Anthony Zhang, Chung~Min Kim, Trevor Darrell, Pieter Abbeel, Jitendra Malik, and Angjoo Kanazawa.
\newblock Visual imitation enables contextual humanoid control.
\newblock \emph{arXiv preprint arXiv:2505.03729}, 2025.

\bibitem[He et~al.(2025{\natexlab{a}})He, Gao, Xiao, Zhang, Wang, Wang, Luo, He, Sobh, Pan, Yi, Qu, Kitani, Hodgins, Fan, Zhu, Liu, and Shi]{He2025ASAP}
Tairan He, Jiawei Gao, Wenli Xiao, Yuanhang Zhang, Zi~Wang, Jiashun Wang, Zhengyi Luo, Guanqi He, Nikhil Sobh, Chaoyi Pan, Zeji Yi, Guannan Qu, Kris~M. Kitani, Jessica Hodgins, Linxi~Jim Fan, Yuke Zhu, Changliu Liu, and Guanya Shi.
\newblock {ASAP}: Aligning simulation and real-world physics for learning agile humanoid whole-body skills.
\newblock In \emph{Robotics: Science and Systems (RSS)}, 2025{\natexlab{a}}.
\newblock to appear.

\bibitem[Xie et~al.(2025)Xie, Han, Zheng, Li, Liu, Shi, Zhang, Bai, and Li]{KungfuBot}
Weiji Xie, Jinrui Han, Jiakun Zheng, Huanyu Li, Xinzhe Liu, Jiyuan Shi, Weinan Zhang, Chenjia Bai, and Xuelong Li.
\newblock Kungfubot: Physics-based humanoid whole-body control for learning highly-dynamic skills.
\newblock \emph{arXiv preprint arXiv:2506.12851}, 2025.

\bibitem[Zhang et~al.(2025{\natexlab{b}})Zhang, Zheng, Nai, Hu, Wang, Chen, Lin, Li, Hong, Sreenath, et~al.]{HUB}
Tong Zhang, Boyuan Zheng, Ruiqian Nai, Yingdong Hu, Yen-Jen Wang, Geng Chen, Fanqi Lin, Jiongye Li, Chuye Hong, Koushil Sreenath, et~al.
\newblock Hub: Learning extreme humanoid balance.
\newblock \emph{arXiv preprint arXiv:2505.07294}, 2025{\natexlab{b}}.

\bibitem[Fuchioka et~al.(2023)Fuchioka, Ogata, and Kajita]{Fuchioka2023OptMimic}
Yukiyasu Fuchioka, Tetsuya Ogata, and Shuuji Kajita.
\newblock {Opt-Mimic}: Imitation of optimized trajectories for dynamic quadruped behaviors.
\newblock In \emph{IEEE Int. Conf. on Robotics and Automation (ICRA)}, pages 6726--6732, 2023.
\newblock \doi{10.1109/ICRA48891.2023.10161217}.

\bibitem[Jenelten et~al.(2024)Jenelten, He, Farshidian, and Hutter]{jenelten2024dtc}
Fabian Jenelten, Junzhe He, Farbod Farshidian, and Marco Hutter.
\newblock Dtc: Deep tracking control.
\newblock \emph{Sci. Robotics}, 9\penalty0 (86):\penalty0 eadh5401, 2024.

\bibitem[Bogdanovic et~al.(2022)Bogdanovic, Khadiv, and Righetti]{Bogdanovic2022TOdemo}
Miroslav Bogdanovic, Majid Khadiv, and Ludovic Righetti.
\newblock Model-free reinforcement learning for robust locomotion using demonstrations from trajectory optimization.
\newblock \emph{Front. Robot. AI}, 9:\penalty0 854212, 2022.
\newblock \doi{10.3389/frobt.2022.854212}.

\bibitem[Sleiman et~al.(2024)Sleiman, Mittal, and Hutter]{Sleiman2025GuidedRL}
Jean-Pierre Sleiman, Mayank~K. Mittal, and Marco Hutter.
\newblock Guided reinforcement learning for robust multi-contact loco-manipulation.
\newblock In \emph{Conference on Robot Learning (CoRL)}, volume 270 of \emph{Proceedings of Machine Learning Research}, pages 2210--2247. PMLR, 2024.

\bibitem[He et~al.(2024)He, Luo, He, Xiao, Zhang, Zhang, Kitani, Liu, and Shi]{He2024OmniH2O}
Tairan He, Zhengyi Luo, Xialin He, Wenli Xiao, Chong Zhang, Weinan Zhang, Kris~M. Kitani, Changliu Liu, and Guanya Shi.
\newblock Omnih2o: Universal and dexterous human-to-humanoid whole-body teleoperation and learning.
\newblock In \emph{Conference on Robot Learning (CoRL)}, volume 270 of \emph{Proceedings of Machine Learning Research}, pages 1516--1540. PMLR, 2024.

\bibitem[He et~al.(2025{\natexlab{b}})He, Xiao, Lin, Luo, Xu, Jiang, Kautz, Liu, Shi, Wang, Fan, and Zhu]{He2024HOVER}
Tairan He, Wenli Xiao, Toru Lin, Zhengyi Luo, Zhenjia Xu, Zhenyu Jiang, Jan Kautz, Changliu Liu, Guanya Shi, Xiaolong Wang, Linxi~Jim Fan, and Yuke Zhu.
\newblock {HOVER}: Versatile neural whole-body controller for humanoid robots.
\newblock In \emph{IEEE Int. Conf. on Robotics and Automation (ICRA)}, 2025{\natexlab{b}}.
\newblock to appear.

\bibitem[Chen et~al.(2025)Chen, Ji, Cheng, Peng, Peng, and Wang]{Chen2025GMT}
Zixuan Chen, Mazeyu Ji, Xuxin Cheng, Xuanbin Peng, Xue~Bin Peng, and Xiaolong Wang.
\newblock Gmt: General motion tracking for humanoid whole-body control.
\newblock \emph{arXiv preprint arXiv:2506.14770}, 2025.

\bibitem[Liao et~al.(2025)Liao, Truong, Huang, Tevet, Sreenath, and Liu]{BeyondMimic}
Qiayuan Liao, Takara~E. Truong, Xiaoyu Huang, Guy Tevet, Koushil Sreenath, and C.~Karen Liu.
\newblock Beyondmimic: From motion tracking to versatile humanoid control via guided diffusion.
\newblock \emph{CoRR}, abs/2508.08241, 2025.
\newblock \doi{10.48550/ARXIV.2508.08241}.

\bibitem[Yang et~al.(2025)Yang, Huang, Wu, Kanazawa, Abbeel, Sferrazza, Liu, Duan, and Shi]{OmniRetarget}
Lujie Yang, Xiaoyu Huang, Zhen Wu, Angjoo Kanazawa, Pieter Abbeel, Carmelo Sferrazza, C.~Karen Liu, Rocky Duan, and Guanya Shi.
\newblock Omniretarget: Interaction-preserving data generation for humanoid whole-body loco-manipulation and scene interaction.
\newblock \emph{CoRR}, abs/2509.26633, 2025.
\newblock \doi{10.48550/ARXIV.2509.26633}.

\bibitem[Luo et~al.(2025)Luo, Yuan, Wang, Li, Chen, Casta{\~{n}}eda, Cao, Li, Minor, Ben, Da, Ding, Hogg, Song, Lim, Jeong, He, Xue, Xiao, Wang, Yuen, Kautz, Chang, Iqbal, Fan, and Zhu]{Sonic}
Zhengyi Luo, Ye~Yuan, Tingwu Wang, Chenran Li, Sirui Chen, Fernando Casta{\~{n}}eda, Zi{-}Ang Cao, Jiefeng Li, David Minor, Qingwei Ben, Xingye Da, Runyu Ding, Cyrus Hogg, Lina Song, Edy Lim, Eugene Jeong, Tairan He, Haoru Xue, Wenli Xiao, Zi~Wang, Simon Yuen, Jan Kautz, Yan Chang, Umar Iqbal, Linxi~Jim Fan, and Yuke Zhu.
\newblock {SONIC:} supersizing motion tracking for natural humanoid whole-body control.
\newblock \emph{CoRR}, abs/2511.07820, 2025.
\newblock \doi{10.48550/ARXIV.2511.07820}.

\bibitem[Tessler et~al.(2024)Tessler, Guo, Nabati, Chechik, and Peng]{Tessler2024MaskedMimic}
Chen Tessler, Yunrong Guo, Ofir Nabati, Gal Chechik, and Xue~Bin Peng.
\newblock Maskedmimic: Unified physics-based character control through masked motion inpainting.
\newblock \emph{ACM Trans. Graph. (Proc. SIGGRAPH Asia)}, 43\penalty0 (6), 2024.
\newblock \doi{10.1145/3687951}.

\bibitem[Mittal et~al.(2025)Mittal, Roth, Tigue, Richard, Zhang, Du, Serrano-Mu{\~n}oz, Yao, Zurbr{\"u}gg, Rudin, et~al.]{mittal2025isaac}
Mayank Mittal, Pascal Roth, James Tigue, Antoine Richard, Octi Zhang, Peter Du, Antonio Serrano-Mu{\~n}oz, Xinjie Yao, Ren{\'e} Zurbr{\"u}gg, Nikita Rudin, et~al.
\newblock Isaac lab: A {GPU}-accelerated simulation framework for multi-modal robot learning.
\newblock \emph{arXiv preprint arXiv:2511.04831}, 2025.

\bibitem[Miller et~al.(2025{\natexlab{b}})Miller, Yu, Brauckmann, and Farshidian]{miller2025high}
AJ~Miller, Fangzhou Yu, Michael Brauckmann, and Farbod Farshidian.
\newblock High-performance reinforcement learning on spot: Optimizing simulation parameters with distributional measures.
\newblock \emph{arXiv preprint arXiv:2504.17857}, 2025{\natexlab{b}}.

\bibitem[Inc.(2025)]{xsens_mocap}
Movella Inc.
\newblock Xsens link motion capture system.
\newblock \url{https://www.movella.com/motion-capture/xsens-mvn-link}, 2025.
\newblock Inertial motion capture hardware and software system.

\bibitem[Ltd.(2025)]{vicon_mocap}
Vicon Motion~Systems Ltd.
\newblock Vicon motion capture system.
\newblock \url{https://www.vicon.com/}, 2025.
\newblock Optical motion capture hardware and software system.

\bibitem[Li et~al.(2025)Li, Tucker, Cole, Wang, Jin, Ye, Kanazawa, Holynski, and Snavely]{li2025megasam}
Zhengqi Li, Richard Tucker, Forrester Cole, Qianqian Wang, Linyi Jin, Vickie Ye, Angjoo Kanazawa, Aleksander Holynski, and Noah Snavely.
\newblock Megasam: Accurate, fast and robust structure and motion from casual dynamic videos.
\newblock In \emph{IEEE/CVF Conf. on Computer Vision and Pattern Recognition (CVPR)}, pages 10486--10496, 2025.

\bibitem[Wang et~al.(2024)Wang, Wang, Liu, and Daniilidis]{wang2024tram}
Yufu Wang, Ziyun Wang, Lingjie Liu, and Kostas Daniilidis.
\newblock Tram: Global trajectory and motion of 3d humans from in-the-wild videos.
\newblock In \emph{European Conference on Computer Vision}, pages 467--487. Springer, 2024.

\bibitem[Gleicher(1998)]{Gle98}
Michael Gleicher.
\newblock Retargeting motion to new characters.
\newblock In \emph{Proc. of ACM SIGGRAPH 98}, Annual Conference Series, pages 33--42. ACM SIGGRAPH, jul 1998.
\newblock \doi{http://dx.doi.org/10.1145/280814.280820}.

\bibitem[Schulman et~al.(2017)Schulman, Wolski, Dhariwal, Radford, and Klimov]{PPO}
John Schulman, Filip Wolski, Prafulla Dhariwal, Alec Radford, and Oleg Klimov.
\newblock Proximal policy optimization algorithms.
\newblock \emph{arXiv preprint arXiv:1707.06347}, 2017.

\bibitem[Pinto et~al.(2017)Pinto, Andrychowicz, Welinder, Zaremba, and Abbeel]{AsymmetricActorCritic}
Lerrel Pinto, Marcin Andrychowicz, Peter Welinder, Wojciech Zaremba, and Pieter Abbeel.
\newblock Asymmetric actor critic for image-based robot learning.
\newblock \emph{arXiv preprint arXiv:1710.06542}, 2017.

\bibitem[Karpathy and van~de Panne(2012)]{Karpathy2012Curriculum}
Andrej Karpathy and Michiel van~de Panne.
\newblock Curriculum learning for motor skills.
\newblock In \emph{Advances in Artificial Intelligence (Canadian AI 2012)}, volume 7310 of \emph{Lecture Notes in Computer Science}, pages 325--330. Springer, 2012.
\newblock \doi{10.1007/978-3-642-30353-1_31}.

\bibitem[Yu et~al.(2018)Yu, Turk, and Liu]{Yu2018Symmetric}
Wenhao Yu, Greg Turk, and C.~Karen Liu.
\newblock Learning symmetric and low-energy locomotion.
\newblock \emph{ACM Trans. Graph.}, 37\penalty0 (4):\penalty0 144:1--144:12, 2018.
\newblock \doi{10.1145/3197517.3201397}.

\bibitem[Huang et~al.(2025)Huang, Ren, Wang, Wang, Ben, Wen, Chen, Li, and Pang]{HoST2025}
Tao Huang, Junli Ren, Huayi Wang, Zirui Wang, Qingwei Ben, Muning Wen, Xiao Chen, Jianan Li, and Jiangmiao Pang.
\newblock Learning humanoid standing-up control across diverse postures.
\newblock In \emph{Robotics: Science and Systems (RSS)}, 2025.
\newblock to appear.

\bibitem[Cao et~al.(2025)Cao, Zhang, Nie, Lin, Li, and Gao]{Cao2025A2CF}
Zhanxiang Cao, Yang Zhang, Buqing Nie, Huangxuan Lin, Haoyang Li, and Yue Gao.
\newblock Learning motion skills with adaptive assistive curriculum force in humanoid robots.
\newblock \emph{arXiv preprint arXiv:2506.23125}, 2025.

\bibitem[Todorov et~al.(2012)Todorov, Erez, and Tassa]{todorov2012mujoco}
Emanuel Todorov, Tom Erez, and Yuval Tassa.
\newblock {MuJoCo}: A physics engine for model-based control.
\newblock In \emph{2012 IEEE/RSJ international conference on intelligent robots and systems}, pages 5026--5033. IEEE, 2012.

\end{thebibliography}

\clearpage
\newpage

\setcounter{table}{0}
\makeatletter 
\renewcommand{\thetable}{S\@arabic\c@table}
\makeatother

\setcounter{figure}{0}
\makeatletter 
\renewcommand{\thefigure}{S\@arabic\c@figure}
\makeatother

\setcounter{algorithm}{0}
\makeatletter 
\renewcommand{\thealgorithm}{S\@arabic\c@algorithm}
\makeatother

\renewcommand{\thefigure}{S\arabic{figure}}
\renewcommand{\thetable}{S\arabic{table}}
\newcommand{\btau}{\boldsymbol{\tau}}
\newcommand{\bGamma}{\boldsymbol{\Gamma}}
\renewcommand{\v}{\mathbf{v}}
\renewcommand{\S}{\mathbf{S}}
\newcommand{\hli}[1]{{\color{blue}#1}}

\clearpage
\onecolumn
{\Huge \bfseries Supplementary Material\par}
\subsection*{Section S1. Nomenclature}
\begin{table}[H]
\centering
\small
\setlength{\tabcolsep}{6pt}
\begin{tabular}{cl}
$I$ & Inertial (world) frame \\
$B$ & Base frame \\
$T$ & Torso frame \\
$j$ & Joint index \\
$kb$ & Keybodies \\
$\mathcal{J}$ & Set of joint indices \\
$n_j$ & Number of joints \\
$n_{kb}$ & Number of keybodies \\
$n_a$ & Number of actions \\[0.35em]

$\mathbf{q}$ & Joint positions vector \\
$\dot{\mathbf{q}}$ & Joint velocities vector \\
$\ddot{\mathbf{q}}$ & Joint accelerations vector \\
$\boldsymbol{\tau}$ & Joint torques vector \\
$\mathbf{M}$ & Mass matrix \\
$\mathbf{h}$ & Nonlinear forces vector \\
$\mathbf{J}$ & Jacobian matrix \\
$\mathbf{F}$ & Force vector \\
$\mathbf{W}$ & Wrench vector \\
$\mathbf{I}$ & (Base) inertia matrix \\[0.35em]

$\mathcal{M}$ & Markov Decision Process \\
$s_t$ & State at time $t$ \\
$a_t$ & Action at time $t$ \\
$o_t$ & Observation at time $t$ \\
$\pi_\theta$ & Policy parameterized by $\theta$ \\
$\theta$ & Policy parameters \\
$T$ & Finite horizon length \\[0.35em]

$N$ & Number of reference trajectories \\
$i\in\{1,\dots,N\}$ & Trajectory index \\
$D_i$ & Continuous duration of trajectory $i$ \\
$L_i$ & Discrete length of trajectory $i$ \\
$\Delta$ & Temporal bin width \\
$B$ & Total number of bins \\
$b\in\{0,\dots,B{-}1\}$ & Bin index \\
$M_{i,b}$ & Validity mask for trajectory $i$ in bin $b$ \\
$\Omega$ & Feasible set of valid trajectory–bin pairs \\
$|\Omega|$ & Cardinality of feasible set of bins \\
$f_{i,b}$ & Failure level for trajectory $i$ in bin $b$ \\
$\alpha$ & EMA smoothing factor \\
$\tau$ & Sampling temperature \\
$\tau_{\text{base}}$ & Base temperature \\
$\varepsilon$ & Uniform floor weight (per-bin lower bound $\ge \varepsilon/|\Omega|$) \\
$t_{\text{init}}$ & Initial time for episode \\
$\phi$ & Normalized phase $\in[0,1]$ \\
\end{tabular}
\end{table}

\begin{table}[H]
\centering
\small
\setlength{\tabcolsep}{6pt}
\begin{tabular}{cl}
$\phi_{\text{init}}$ & Initial phase for episode reset \\
$\ell_{i,b}$ & Logits for sampling \\
$p_{i,b}$ & Sampling probabilities \\
$\beta_{i,b}$ & Assistance scale for trajectory $i$ in bin $b$ \\
$\beta_{\max}$ & Maximum assistance scale \\
$\beta_e$ & Environment-specific assistance scale \\
$\hat{S}_{i,b}$ & Smoothed similarity metric \\
$\eta$ & Target similarity threshold \\
$\mathbf{F}_b$ & Assistive force vector \\
$\mathbf{M}_b$ & Assistive moment vector \\
$\mathbf{w}_e$ & Applied assistive wrench \\
$k_p^v,\,k_d^v$ & Virtual force PD gains \\
$k_p^\omega,\,k_d^\omega$ & Virtual torque PD gains \\
$\mathbf{r}_{\mathrm{b,com}}$ & CoM position w.r.t. base \\[0.35em]

$s_k$ & Instantaneous similarity metric at step $k$ \\
$\bar{s}_e$ & Length-normalized similarity metric for episode $e$ \\
$L_{\mathrm{real}}$ & Realized episode length \\
$L_{\max}$ & Maximum possible episode length \\
$\bar{L}_{\mathrm{episode}}$ & Episode length limit \\
$r_{\text{total}}$ & Total reward \\
$r_{\text{track}}$ & Tracking reward component \\
$r_{\text{reg}}$ & Regularization reward component \\
$r_{\text{survival}}$ & Survival reward component \\
$c_{t_i}$ & Tracking reward weight for term $i$ \\
$\mathbf{e}_i$ & Error vector for tracking term $i$ \\
$\sigma$ & Standard deviation for tracking rewards \\
$\kappa$ & Stiffness parameter for tracking rewards \\
$dt$ & Control timestep \\[0.35em]

$\mathbf{p},\,\mathbf{v}$ & Position and linear velocity \\
$\boldsymbol{\Phi},\,\boldsymbol{\omega}$ & Orientation and angular velocity \\
$\mathbf{g}$ & Gravity vector \\
$\mathbf{r}_{XY}$ & Position of frame $Y$ w.r.t. frame $X$ \\
$\mathbf{v}_{XY}$ & Velocity of frame $Y$ w.r.t. frame $X$ \\
$\boldsymbol{\omega}_{XY}$ & Angular velocity of frame $Y$ w.r.t. frame $X$ \\
${}_X(\cdot)$ & Quantity expressed in frame $X$ \\
$\hat{(\cdot)}$ & Reference (desired) value \\
$(\cdot)^*$ & Reference (desired) value (alternative notation) \\[0.35em]

$\mathbb{E}[\cdot]$ & Expectation operator \\
$\|\cdot\|$ & Euclidean norm \\
$\exp(\cdot)$ & Exponential function \\
$\mathrm{clip}(\cdot)$ & Clipping function \\
$\mathcal{U}(a,b)$ & Uniform distribution between $a$ and $b$ \\
$\mathcal{N}(\mu,\sigma^2)$ & Normal distribution \\
$\mathbf{1}\{\cdot\}$ & Indicator function \\
$\lceil\cdot\rceil$ & Ceiling function \\
$\boxminus$ & Lie-group difference on $SO(3)$ \\
$\times$ & Cross product operator \\
\end{tabular}
\end{table}

\newpage
\subsection*{Section S2. Implementation Details: Actuator Modeling}
\label{sec:supplementary_actuator}


The PLA systems in humanoid robots are essential hardware components that help reduce limb inertia by positioning the actuators closer to the body, which enhances dynamic performance. Nonetheless, enforcing their loop-closure constraints leads to large internal forces, making the simulation of dynamics stiff and substantially increasing the computational cost of the iterative Temporal Gauss-Seidel solver used for numerical integration. Since an entirely accurate model is too slow for large-scale RL training, we tackle this trade-off between fidelity and efficiency by developing a series of progressive approximations to model the PLAs.
This section details the formulation of our progressive PLA approximation models and provides an experimental validation of their accuracy. It concludes the section by showcasing the importance of considering configuration-dependent torque limits of the PLA actuators.

\subsubsection*{Projected Model}
\begin{figure}[!ht]
    \centering
    \includegraphics[width=0.7\linewidth]{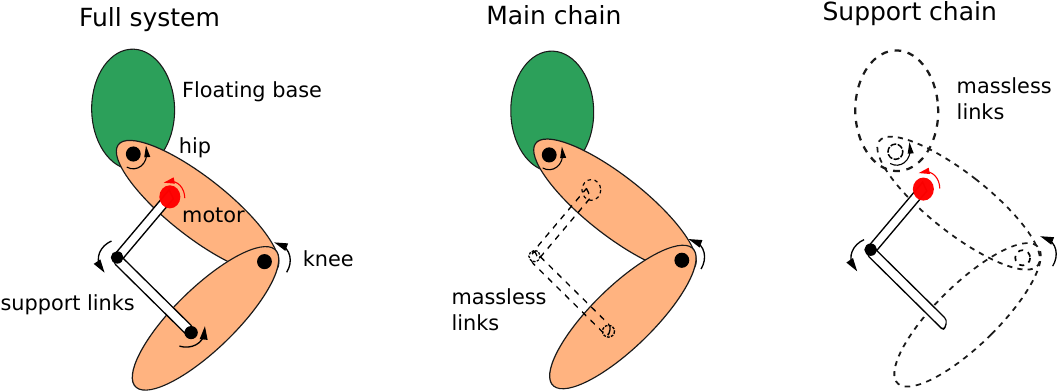}
    \caption{Illustration of main chain and support chain. Dashed links are assumed to be massless.}
    \label{fig:illustration_main_support}
\end{figure}

Here, we formulate an analytically exact \emph{Projected Model} to circumvent the numerical constraint enforcement used in standard simulators. We begin by deriving the general equation of motion for a class of systems that contain a closed-kinematic chain. An instance of such a mechanism is the four-bar linkage, illustrated in Figure 4, where the primary kinematic chain consists of the floating base, thigh, shank, and their connecting joints. The PLA's actuation is provided by a motor mounted on the thigh, which drives the knee joint through a four-bar linkage. Our objective is to project the dynamics of the PLA's support links onto the main kinematic chain. To analyze the dynamics of PLA, we conceptually decompose its closed-loop structure into two overlapping open kinematic chains: a \emph{main chain}, $\mathcal{M}$, and a \emph{support chain}, $\mathcal{S}$, as illustrated in Figure.\ref{fig:illustration_main_support}.
\begin{itemize}
    \item The \emph{main chain}, $\mathcal{M}$, is comprised of the robot's primary kinematic tree, the PLA's parent link, and its unactuated branch.
    \item The \emph{support chain}, $\mathcal{S}$, consists of the PLA's support links, which also originate from the same parent link.
\end{itemize}
A central assumption of this decomposition is that while the parent link is fully modeled with its physical mass in chain $\mathcal{M}$, it is considered massless in the support chain, $\mathcal{S}$, to avoid double-counting inertial properties. We will further assume that the velocity of the PLA joints in the support kinematic chain, including the actuated joints $\q_i$ and dependent joints, $\q_d$, can be calculated based on the velocity of the PLA joints in the main kinematic chain, $\q_o$, 
\begin{equation}
    \mat{\Dot{\q}_d \\ \Dot{\q}_i} = \mat{\boldsymbol{\Gamma}_d(\q_o) \\ \boldsymbol{\Gamma}_i(\q_o)} \dot{\q}_o,
    \label{eq:out_to_support}
\end{equation}
where $\boldsymbol{\Gamma}_d$ and $\boldsymbol{\Gamma}_i$ denote the kinematic mappings from the main chain joint velocities to the dependent and actuated joints, respectively.

To simplify our derivation, we focus on a scenario where the parent joints are unactuated and the entire system is free from external forces, with actuation occurring only through the PLA. While extending this analysis to a more general case is straightforward, we will not cover it here to keep the discussion concise. The dynamics of the main chain and auxiliary chain can then be expressed as follows:
\begin{align}
    \label{eq:main_eom}
    \M_{\mathcal{M}} &\mat{\ddot{\q}_p \\ \ddot{\q}_o} + \h_{\mathcal{M}}  = \J_{\mathcal{M}}^{\top} \F \\
    \label{eq:support_eom}
    \M_{\mathcal{S}} &\mat{\ddot{\q}_p \\ \ddot{\q}_d \\ \ddot{\q}_i} + \h_{\mathcal{S}}  = \mat{ 0 \\ 0 \\ \btau_i } - \J_{\mathcal{S}}^{\top} \F,
\end{align}
where $\M_{\mathcal{M}}$ and $\h_{\mathcal{M}}$ are the mass matrix and the nonlinear forces of the main chain, and $\M_{\mathcal{S}}$, $\h_{\mathcal{S}}$ denote those of the support chain. The variable $\q_p$ corresponds to the parent joints, and $\btau_i$ represents the torque applied at the PLA's actuated joint. $\F$ is the generalized interaction force exchanged between the main and support chains at their connection point, and $\J_{\mathcal{M}}$ and $\J_{\mathcal{S}}$ are the corresponding Jacobians at that point. Given that the velocity of the point should be equal, we have
\begin{equation*}
    \J_{\mathcal{S}} \mat{\dot{\q}_p \\ \dot{\q}_d \\ \Dot{\q}_i} = \J_{\mathcal{M}} \mat{\dot{\q}_p \\ \dot{\q}_o}.
\end{equation*}

Applying the mapping in~\ref{eq:out_to_support} yields the corresponding relationship between the Jacobians
\begin{equation}
    \label{eq:g_map}
    \J_{\mathcal{M}} = \J_{\mathcal{S}} \underbrace{\mat{\I & 0 \\ 0 & \boldsymbol{\Gamma}_d(\q_o) \\ 0 & \boldsymbol{\Gamma}_i(\q_o)}}_{\G(\q_o)}.
\end{equation}

Based on \ref{eq:support_eom}, \ref{eq:main_eom}, and \ref{eq:g_map}, the dynamics of the \textit{Projected} Model can be derived as
\begin{equation}
    \big(\M_{\mathcal{M}} + \G^{\top}\M_{\mathcal{S}}\G\big)\mat{\ddot{\q}_p \\ \ddot{\q}_o}
         + \h_{\mathcal{M}} + \G^{\top}\h_{\mathcal{S}} + \G^{\top}\M_{\mathcal{S}}\Dot{\G} \mat{0 \\ \dot{\q}_o} = \mat{0 \\ \btau_o},
\end{equation}
where $\btau_o$ denotes the output torque of PLA. The transmission Jacobian $\boldsymbol{\Gamma}_i$ defines the relationship between the input torque and the output torque of the PLA, expressed as:
\begin{equation}
    \label{eq:torque_map}
    \btau_o = \boldsymbol{\Gamma}_i^\top(\q_o) \, \btau_i.
\end{equation}

Although this model is precise, its governing equations do not adhere to the standard rigid body structure required by off-the-shelf physics engines. This structural incompatibility prevents its direct application, leading us to create subsequent approximate models that are tailored to be compatible with simulators.

\subsubsection*{Locally Projected Model: Massless-Links Approximation}
Our first approximation is motivated by the lightweight design of the support links in the parallel kinematic chain of PLAs. We assume that these support links are massless while still accounting for the full inertia of the motor armatures and the main kinematic chain of PLA.  By treating the support links as massless and considering the Coriolis forces, \( \h_{\mathcal{S}} \), to be negligible, we can derive:
\begin{equation}\label{eq:LPM}
    \Big(\M_\mathcal{M} + \mat{0 & 0 \\ 0 & \M_o} \Big) \mat{\ddot{\q}_p \\ \ddot{\q}_o}
         + \h_\mathcal{M} + \mat{0 \\ \h_0} = \mat{0 \\ \btau_o} ,
\end{equation}
where $\M_o(\q_o) = \boldsymbol{\Gamma}_i^\top \I_i \boldsymbol{\Gamma}_i$ is the locally projected mass matrix, under the assumption that $\M_\mathcal{S} = \text{Diag}(0, \I_i)$ is diagonal and only encodes the PLA actuator armature values, with $\I_i$ representing the motor armatures of the PLA. The term $\h_0(\q_o, \dot{\q}_o) = \boldsymbol{\Gamma}_i^\top \I_i \dot{\boldsymbol{\Gamma}}_i \dot{\q}_o$ represents the vector of nonlinear forces arising from the PLA dynamics, which acts locally on the PLA's output joints.

This model results in a valid rigid body model that is compatible with standard simulators. In this model, the armature values (inertia terms added to the diagonal of the mass matrix) depend on the configuration. In a simulator like Isaac Lab, this can be implemented by updating the armature values at each simulation step using a lookup table.

\subsubsection*{Dynamic Armature Model: Diagonal Approximation}
Although the \emph{Locally Projected Model} is local, meaning that the dynamics of the PLA actuators only affect themselves and are solely a function of the PLA's output joint, it should be noted that $\M_o$ is not necessarily diagonal even though $\I_\mathcal{i}$ is diagonal. Since standard simulators cannot accommodate off-diagonal terms, this model faces limitations when applied to coupled PLAs, such as those found in the ankles of Atlas and G1, where two actuators simultaneously drive the ankle pitch and roll joints. To resolve this, we propose our second approximation, which we refer to as the \emph{Dynamic Armature Model}. To derive this model, we decompose $\M_o$ into two components: $D_{o}$, which is a diagonal matrix, and $O_{o}$, which represents the off-diagonal elements. We then define the model as follows:
\begin{equation}\label{eq:DAM}
    \Big(\M_\mathcal{M} + \mat{0 & 0 \\ 0 & D_o} \Big) \mat{\ddot{\q}_p \\ \ddot{\q}_o}
         + \h_\mathcal{M} + \mat{0 \\ \h_0 + O_o \ddot{\q}_{o}} = \mat{0 \\ \btau_o} .
\end{equation}

In the next step, instead of using $\ddot{\q}_{o}$, we will use the previous joint acceleration of the PLA, denoted as $\ddot{\q}_{o}^{'}$.
\begin{equation}\label{eq:DAM}
    \Big(\M_\mathcal{M} + \mat{0 & 0 \\ 0 & D_o} \Big) \mat{\ddot{\q}_p \\ \ddot{\q}_o}
         + \h_\mathcal{M} + \mat{0 \\ \h_0 + O_o \ddot{\q}_{o}^{'}} = \mat{0 \\ \btau_o} .
\end{equation}

The \emph{Dynamic Armature Model} provides a Jacobi approximation of the \emph{Locally Projected Model}, representing the PLA dynamics as two components: a configuration-dependent diagonal armature and a fictitious torque. This fictitious torque, which accounts for the off-diagonal dynamic effects, is calculated using the PLA joint acceleration from the previous timestep. This approximation introduces only a transient error, provided the system's mass matrix is diagonally dominant. We have empirically verified that this condition holds for the Atlas ankle by sweeping its whole workspace.

\subsubsection*{Nominal Armature Model: Fixed-Configuration Approximation}
Thus far, our approximations have produced a compatible model with minimal error. However, the suggested configuration-dependent armature matrix creates a practical bottleneck. Updating armature values at each simulation step incurs a significant computational overhead (approximately 20\% in Isaac Lab). To address this issue, we introduce our final and strongest approximation, the \emph{Nominal Armature Model}. This model computes the joint armatures once at a single, well-chosen nominal configuration and then fixes these values throughout training, denoted by $\Bar{D}_o$ and $\Bar{O}_o$. The dynamics of the \emph{Nominal Armature Model} are as follows:
\begin{equation}\label{eq:NAM}
    \Big(\M_\mathcal{M} + \mat{0 & 0 \\ 0 & \Bar{D}_o} \Big) \mat{\ddot{\q}_p \\ \ddot{\q}_o}
         + \h_\mathcal{M} + \mat{0 \\ \h_0 + \Bar{O}_o \ddot{\q}_{o}^{'}} = \mat{0 \\ \btau_o} ,
\end{equation}

This model enables adding the projected armature (of the motors on the closed kinematic chain) to the main kinematic tree when training is initialized. Crucially, this single set of nominal armature values serves a dual purpose: it simplifies the simulation dynamics and it provides a principled basis for designing the fixed PD gains of our low-level controller.

\subsubsection*{PLA Models Evaluation}
To quantify the accuracy of our approximations, we compare them to the actual model. In addition to the models mentioned above, we also evaluate a baseline known as the \emph{Simplest Model}, which excludes all fictitious torque from the \emph{Nominal Armature} model. 

The tests are conducted using the Atlas ankle, which is commanded to follow a 5~Hz sinusoidal trajectory in both pitch and roll, with an amplitude of half the joint position limit. To measure performance, we calculate the joint acceleration of each approximate model and determine the error relative to the actual model. This error is normalized at each timestep, and we report the final Mean Squared Error (MSE) for each model in Table \ref{tab:actuator_model_approx}.

The \emph{Locally Projected}, \emph{Dynamic Armature} and \emph{Nominal Armature} models all demonstrate behaviors that closely match those of the exact model, indicating that the assumption of massless support links and the Jacobi approximation are valid. Although the \emph{Simplest Model} shows relatively larger errors, these errors primarily manifest as a shift in the motion, and the resulting motion is quite similar to that of the actual model. Given that the motion of the \emph{Simplest Model} is qualitatively similar, its simplicity makes it an attractive option, especially since the RL policy could learn to compensate for these model discrepancies. As a result, in some of the training runs, we opted to use this model.

\subsubsection*{Impact of PLA Modeling on Sim-To-Real Transfer}
To better isolate the effect of PLA modeling on sim-to-real transfer, we performed a controlled comparison that holds the remainder of the pipeline fixed while varying only the armature model, which is also used to compute the joint-level PD gains. Specifically, we compare our \emph{nominal armature model} for the Unitree G1 robot, which accounts for the parallel linkages at the waist and ankle, against an \emph{unprojected} variant that ignores these linkages. For both models, we use the same PD gain design recipe and hyperparameters (i.e., the same natural frequency and damping ratio), and we keep all other training components identical. In simulation, the two setups exhibit similar training curves and comparable success during rollout evaluation. On hardware, however, the difference becomes clear: across three representative behaviors---namely the cartwheel motion, the ping-pong sequence, and the box climb down---the policy using the nominal armature model transfers reliably, whereas the policy trained with unprojected armatures consistently fails during execution. Qualitative comparisons are provided in Movie~S8.

\subsubsection*{PLA Torque Limits}
\begin{figure}[!ht]
    \centering
    \includegraphics[width=0.7\linewidth]{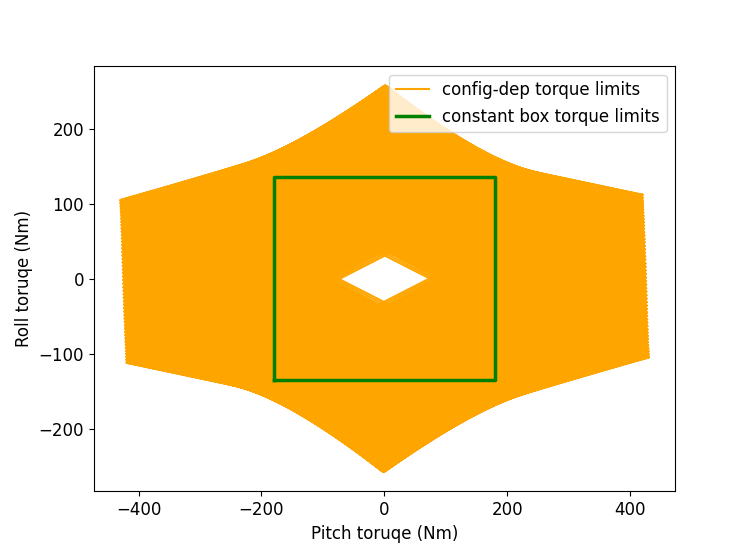}
    \caption{Configuration-dependent torque limit versus box torque limit for Atlas ankle.}
    \label{fig:ankle_torque_limit}
\end{figure}

The equation \eqref{eq:torque_map} defines the relationship between the input torque and the output torque of the PLA. This mapping is dependent on the configuration, meaning that the output torque limits are not simply box limits, even though the input space does have box limits. An approximation approach is to use box limits. However, this can lead to either overly conservative or overly optimistic results. Figure \ref{fig:ankle_torque_limit} illustrates the configuration-dependent torque limits on the Atlas ankle at specific joint roll and pitch angles, compared to the fixed boundary approximation. We sweep the ankle pitch–roll joint space over its position limits. At each pitch–roll joint pair, we calculate the transmission Jacobian $\bGamma_i$ (eq.\eqref{eq:out_to_support}) and apply this linear transformation to the box limit in the motor space, resulting in a sequence of parallelograms. As shown, the fixed limits significantly underestimate the possible torque output.


\subsection*{Section S3. Implementation Details: Spot Modeling}
\label{sec:supplementary_spot}

To bridge the sim-to-real gap for the more challenging continuous backflip behavior, we incorporated a more accurate model of Spot’s power system and actuators in simulation. 
First, we modeled Spot's power system. Spot employs a power-limiting algorithm that sums the requested mechanical power and resistive losses across all actuators and selectively saturates the commanded torque to maintain the total system power within defined limits. We replicated this algorithm in simulation so that policies would experience the same constraints during training~\footnote{Further details of this algorithm cannot be provided, as they involve proprietary industrial information.}. Second, we modeled the motor and transmission efficiencies. At high currents, the motor constant of an electric motor decreases. We modeled this magnet saturation effect using the following equation:
\begin{equation*}
    \tau_{out} = \frac{\tau_{in}}{1+k|\tau_{in}|},
\end{equation*}
with $\tau_{out}$ denoting output torque, $\tau_{in}$ input torque, and $k$ a derating constant. Torque losses from transmission inefficiencies and friction were modeled using the following equation:
\begin{equation*}
    \tau_{out} = \eta(\tau_{in}-I\alpha)-K_c\tanh{(s\omega)}-K_v\omega,
\end{equation*}
where $I$ is the rotor inertia, $\alpha$ the joint acceleration, $K_c$ the Coulomb friction constant, $s$ a smoothing factor, $\omega$ the joint velocity, $K_v$ the viscous friction constant, and $\eta$ the transmission efficiency. Two values of $\eta$ are used, depending on whether the motor performs positive or negative work. The complete actuator model is thus expressed as:
\begin{equation*}
    \tau_{out} = \eta \Big(\frac{\tau_{in}}{1+k|\tau_{in}|}-I\alpha \Big)-K_c\tanh{(s\omega)}-K_v\omega.
\end{equation*}

The actuator model output was low-pass filtered for simulation stability. Parameters $k$, $I$, $K_c$, $s$, and $K_v$ were sourced directly from the manufacturer's specifications. To define the randomization range for $\eta$, we recorded joint torques during dynamic maneuvers (such as a continuous backflip) and estimated the parameters through optimization. The minimum and maximum values set the domain randomization ranges for each actuator type. 

\subsection*{Section S4. Implementation Details: Motion Dataset}
\label{sec:supplementary_data}
In this section, we briefly present the three data sources used in this project.

\subsubsection*{MoCap Data}
MoCap was used to gather high-quality motion references of a human using a combination of Xsens and Vicon systems. Human skeletons from different data sources can vary in animation bone layout, naming, and proportions due to differences in software conventions and variations in morphology among actors. Retargeting is performed in two stages to convert these into robot trajectories.
In the first stage, animation of skeletons is exported to \emph{.bvh} format and retargeted to a target skeleton with human-like kinematics and spherical joints, but proportions similar to the target robot. Alignment frames are placed on the source skeleton bones corresponding to the head, torso, upper arms, lower arms, hands, pelvis, knees, and feet. Optimization is used to minimize spacetime constraint violations between those frames and their counterparts on the target skeleton by solving for the target skeleton's joint poses at all timesteps. We also jointly optimize a uniform scaling of the source data, resampling time inversely with the scale to preserve a gravitational constant of $9.81 \,\text{m/s}^2$ in ballistic motion. The second stage applies kinematic retargeting, mapping the resulting motions to the target robot.

\subsubsection*{ViCap Data}
To leverage the vast amount of motion data available in online videos, we use a custom 3D motion extraction pipeline to reconstruct global human motion, including full-body poses and joint angles, from monocular video recordings (e.g., from a cellphone). While this process yields detailed, high-fidelity motion, the resulting data is prone to more artifacts than MoCap. The process consists of two main stages. In the first stage, we utilize \emph{MegaSaM} to recover the camera trajectory in the world frame, as well as the underlying scene structure and the global metric scale. In the second stage, we estimate human motion as an \emph{SMPL} model in the camera frame using \emph{TRAM}, and then map these poses to world coordinates using the transformation derived from \emph{MegaSaM}. While the original \emph{TRAM} implementation integrates these two steps, we found that substituting the first stage with \emph{MegaSaM} significantly reduces artifacts such as character misorientation, sliding, and floating. Finally, the extracted \emph{SMPL} motion is retargeted to either the Atlas or G1 through kinematic retargeting. Before this process, the \emph{SMPL} skeleton is uniformly scaled to match the target robot, with the scaling factor chosen by matching the \emph{SMPL} thigh bone length to that of the robot.

\subsubsection*{Keyframe Animation}
We utilize keyframe animation for two primary purposes: first, to create movements that are not achievable by humans, such as utilizing the continuous rotation joint of the robot; and second, to generate motions for non-humanoid morphologies, like the Spot quadruped, where collecting real-world animal data is challenging. As these motions are authored with the target robot's kinematics in mind, they typically do not require any retargeting. 

\subsection*{Section S5. Implementation Details: Adaptive RSI Sampler}

Given a library containing $N$ reference trajectories indexed by $i\!\in\!\{1,\dots,N\}$ with continuous durations $D_i$ and discrete lengths $L_i$. We fix a temporal bin width $\Delta{>}0$ and partition time into bins $b\!\in\!\{0,\dots,B{-}1\}$ where
\[
B \;=\; \left\lceil \max_{i} \frac{D_i}{\Delta} \right\rceil.
\]
A validity mask
\[
M_{i,b}\;=\;\mathbf{1}\!\left\{\,b\cdot\Delta < D_i\,\right\}
\]
marks bins that intersect the support of trajectory $i$. We define the feasible set ${\Omega \!=\! \{(i,b)\,:\,M_{i,b}\!=\!1\}}$ and maintain a per-bin \emph{failure level} matrix $f \in \mathbb{R}^{N\times B}$ on $\Omega$. For numerical convenience, we set $f_{i,b}\!=\!{-}\infty$ for $(i,b)\!\notin\!\Omega$ so that invalid bins receive zero probability under a softmax. For each rollout $e$ that \emph{starts} in bin $(i,b)$, we compute a length-normalized similarity metric that captures overall tracking performance
\[
\bar{s}_e \;=\; \frac{1}{L_{\max}}\,\sum_{k=1}^{L_{\mathrm{real}}} s_k,\qquad s_k\in[0,1],
\]
where $s_k$ is given by the joint tracking reward. $L_{\mathrm{real}}$ is the realized episode length and $L_{\max} = \min(\bar{L}_{\text{episode}}, L_i)$ is the total number of steps that could have been realized when early terminations are not triggered (so early terminations are automatically penalized by missing terms). We then update the failure level with an exponential moving average (EMA):
\[
f_{i,b}\;\leftarrow\;(1-\alpha)\,f_{i,b}\;+\;\alpha\,\bigl(1-\bar{s}_e\bigr), \qquad \alpha\in(0,1).
\]
To sample the next initialization bin, we form logits on valid bins,
\[
\ell_{i,b}\;=\;\frac{f_{i,b}}{\tau},\qquad \tau>0,
\]
and draw from the floor-smoothed categorical distribution
\[
p_{i,b}\;=\;(1-\varepsilon)\,\frac{e^{\ell_{i,b}}}{\sum_{(u,v)\in\Omega}e^{\ell_{u,v}}}\;+\;\varepsilon\,\frac{1}{|\Omega|},\qquad \varepsilon\in(0,1),
\]
with $p_{i,b}\!=\!0$ for $(i,b)\!\notin\!\Omega$. The temperature $\tau$ controls concentration around hard bins; the uniform floor $\varepsilon$ preserves global coverage and mitigates forgetting. Finally, given $(i,b)\!\sim\!p$, we sample a start time for trajectory $i$
\[
t_{init}\;\sim\;\mathcal{U}\!\bigl(b\cdot\Delta,\;\min\{(b{+}1)\Delta,\,D_i\}\bigr),
\]
and convert to a normalized phase $\phi\!=\!t_{init}/D_i\in[0,1]$.

\subsection*{Section S6. Implementation Details: Assistive-Wrench Curriculum}

We convert per-bin failure levels $f_{i,b}$ back to a smoothed similarity metric $\hat{S}_{i,b}\!=\!1-f_{i,b}$ on $\Omega$ and map them to a non-negative assistance scale via a monotone schedule that vanishes at a target similarity~$\eta$:
\[
\beta_{i,b}\;=\;\mathrm{clip}\!\Bigl(1-\tfrac{\hat{S}_{i,b}}{\eta},\;0,\;\beta_{\max}\Bigr),\qquad \eta>0,\; \beta_{\max}\!\in[0,1]\!
\]
At initialization, we set the environment’s assistive gain based on its sampled trajectory-bin pair $(i,b)$ (i.e., $\beta_e = \beta_{i,b}$) and we hold $\beta_e$ constant during that episode.

\paragraph{Virtual wrench computation.}
Let $(\mathbf{p},\mathbf{v},\boldsymbol{\Phi},\boldsymbol{\omega})$ denote the base position, linear velocity, orientation, and angular velocity; a hat on top of each variable denotes its corresponding reference. We compute a nominal spatial wrench acting on the robot's torso as follows:
\begin{subequations}\label{eq:assist}
\begin{align}
\mathbf{F}_{b} \;&=\; M\!\left(\hat{\dot{\mathbf{v}}} \;+\; k_p^v\,(\hat{\mathbf{p}}-\mathbf{p}) \;+\; k_d^v\,(\hat{\mathbf{v}}-\mathbf{v}) \;-\; \mathbf{g}\right), \label{eq:assist-F}\\[2pt]
\mathbf{M}_{b} \;&=\; \mathbf{I}\,\hat{\dot{\boldsymbol{\omega}}}
\;+\; k_p^\omega\,\mathbf{I}\,\bigl(\hat{\boldsymbol{\Phi}}\boxminus \boldsymbol{\Phi}\bigr)
\;+\; k_d^\omega\,\mathbf{I}\,(\hat{\boldsymbol{\omega}}-\boldsymbol{\omega})
\;+\; \boldsymbol{\omega}\times(\mathbf{I}\boldsymbol{\omega})
\;-\; \mathbf{r}_{\mathrm{b,com}}\times M\mathbf{g}, \label{eq:assist-M}
\end{align}
\end{subequations}
where $M$ and $\mathbf{I}$ are the whole-body mass and nominal base inertia at a default configuration, $\mathbf{g}$ is gravity, $\mathbf{r}_{\mathrm{b,com}}$ is the position of the whole-body CoM with respect to the base, and $\boxminus$ denotes the Lie-group difference on $SO(3)$ (implemented via the rotation-log map for quaternions). The applied assistive wrench is
\[
\mathbf{w}_e \;=\; \beta_e \begin{bmatrix}\mathbf{F}_b \\ \mathbf{M}_b\end{bmatrix},
\]
expressed in the world frame and applied at the torso link. The cap $\beta_{\max}{<}1$ keeps assistance partial to encourage meaningful exploration and avoid overfitting to the modified physics.

\paragraph{Coupling to the adaptive RSI sampler.}
Because $w_{i,b}$ is a deterministic function of $\hat{S}_{i,b}$, bins with persistent errors (large $f_{i,b}$, small $\hat{S}_{i,b}$) receive stronger assistance initially. As tracking improves, the EMA drives $f_{i,b}$ down, which in turn reduces $w_{i,b}$ and eventually clamps it to zero once ${\hat{S}_{i,b}\!\ge\!\eta}$. Simultaneously, the adaptive RSI scheme increases the sampling frequency of challenging bins until their failure levels subside while preserving coverage via the sampler’s floor probability, yielding a simple, performance-coupled continuation scheme that is guaranteed to decay assistance to zero at the desired similarity threshold.

\newpage

\begin{figure}[H]
    \centering
    \begin{subfigure}[t]{0.49\linewidth}
        \centering
        \includegraphics[width=\linewidth]{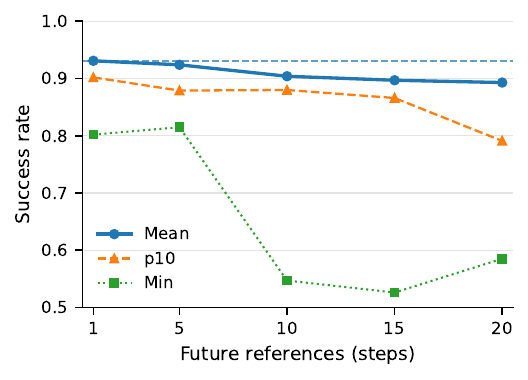}
        \caption{\textbf{Future references.} Baseline uses 1 future reference.}
        \label{fig:abl_future_refs}
    \end{subfigure}\hfill
    \begin{subfigure}[t]{0.49\linewidth}
        \centering
        \includegraphics[width=\linewidth]{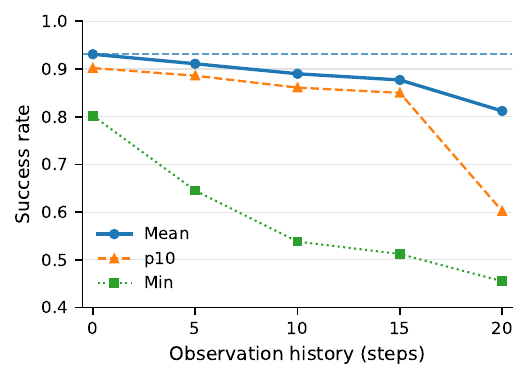}
        \caption{\textbf{Observation history.} Baseline uses 0 history.}
        \label{fig:abl_obs_hist}
    \end{subfigure}

    \caption{\textbf{Effect of future references and observation history on success.}
    We report success statistics as a function of (left) the number of future reference steps
    and (right) the observation history length. Curves show the mean success rate, the 10th percentile (p10),
    and the minimum across multiple trajectories. Overall, increasing the future-reference horizon tends to
    reduce performance, with degraded lower-tail behavior (p10/min) at longer horizons. Increasing observation
    history is even more detrimental: performance drops more sharply, especially in the lower tail, as history
    length grows, indicating that stacking past observations is harmful in this setting.}

    \label{fig:abl_future_and_history}
\end{figure}

\begin{figure}[H]
    \centering

    \begin{subfigure}[t]{0.32\linewidth}
        \centering
        \includegraphics[width=\linewidth]{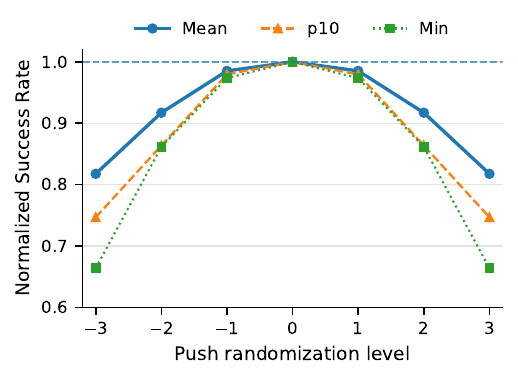}
        \caption{\textbf{Pushes.}}
        \label{fig:rob_push}
    \end{subfigure}\hfill
    \begin{subfigure}[t]{0.32\linewidth}
        \centering
        \includegraphics[width=\linewidth]{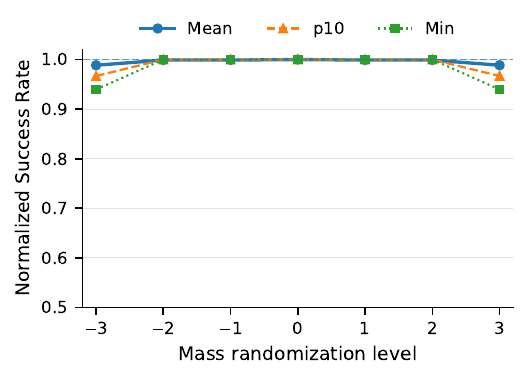}
        \caption{\textbf{Mass.}}
        \label{fig:rob_mass}
    \end{subfigure}\hfill
    \begin{subfigure}[t]{0.32\linewidth}
        \centering
        \includegraphics[width=\linewidth]{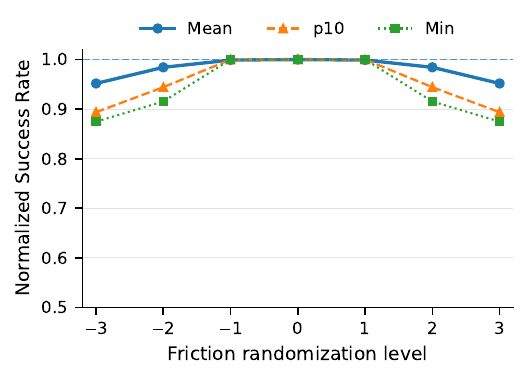}
        \caption{\textbf{Friction.}}
        \label{fig:rob_friction}
    \end{subfigure}

    \vspace{2mm}

    \begin{subfigure}[t]{0.32\linewidth}
        \centering
        \includegraphics[width=\linewidth]{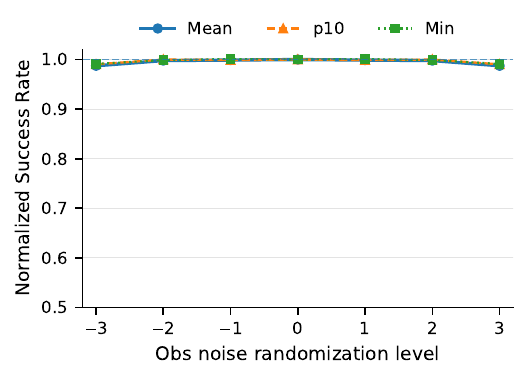}
        \caption{\textbf{Observation noise.}}
        \label{fig:rob_obsnoise}
    \end{subfigure}\hfill
    \begin{subfigure}[t]{0.32\linewidth}
        \centering
        \includegraphics[width=\linewidth]{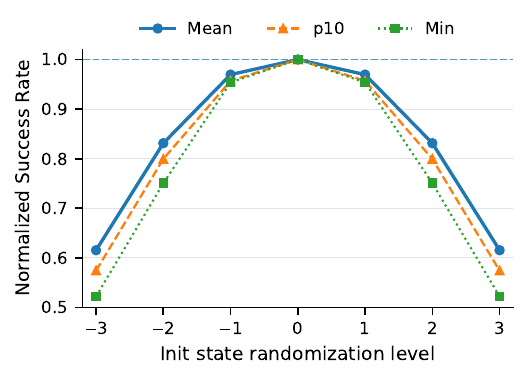}
        \caption{\textbf{Initial state.}}
        \label{fig:rob_init}
    \end{subfigure}\hfill
    \begin{subfigure}[t]{0.32\linewidth}
        \centering
        \includegraphics[width=\linewidth]{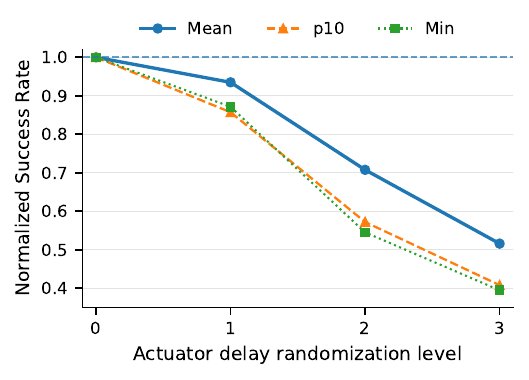}
        \caption{\textbf{Actuator delay.}}
        \label{fig:rob_delay}
    \end{subfigure}

    \caption{\textbf{Robustness to model uncertainty and disturbances.}
    We evaluate a single policy under controlled robustness stress tests and report normalized success statistics
    (mean, p10, and minimum over multiple trajectories), divided by the nominal (no-randomization) performance so that
    the baseline at randomization level 0 equals 1.0. Unless noted otherwise, the policy was trained with
    randomization level 1, and levels 2--3 extrapolate beyond the training distribution.
    \textbf{Actuator delay} has the strongest effect in this suite; importantly, delay randomization was \emph{not} included
    during training, yet performance remains acceptable for modest delays (randomization levels 1--3, corresponding to
    uniformly sampled delays of 0--1, 0--2, and 0--3 control steps), suggesting that other training randomizations
    (e.g., observation/action noise and dynamics randomization) provide indirect robustness to latency.
    Among the randomizations used during training, \textbf{initial-state perturbations} have the largest impact, producing
    the steepest drop in both mean and lower-tail performance. We attribute this primarily to the fact that aggressive
    initial-state perturbations can spawn the robot in self-colliding configurations (e.g., arm--torso collisions) that are
    effectively irreversible under the controller, leading to a higher probability of early failure.
    \textbf{Push disturbances} and \textbf{friction variation} exhibit the next strongest degradation, particularly in the tail
    (p10/min), reflecting reduced robustness under aggressive external forces and contact parameter shifts.
    In contrast, \textbf{mass variation} and \textbf{observation noise} have comparatively mild effects over the tested range,
    suggesting the learned controller is largely invariant to these sources.}

    \label{fig:robustness_suite}
\end{figure}

\begin{figure}[H]
  \centering
  \begin{subfigure}[t]{0.49\columnwidth}
    \centering
    \includegraphics[width=\columnwidth]{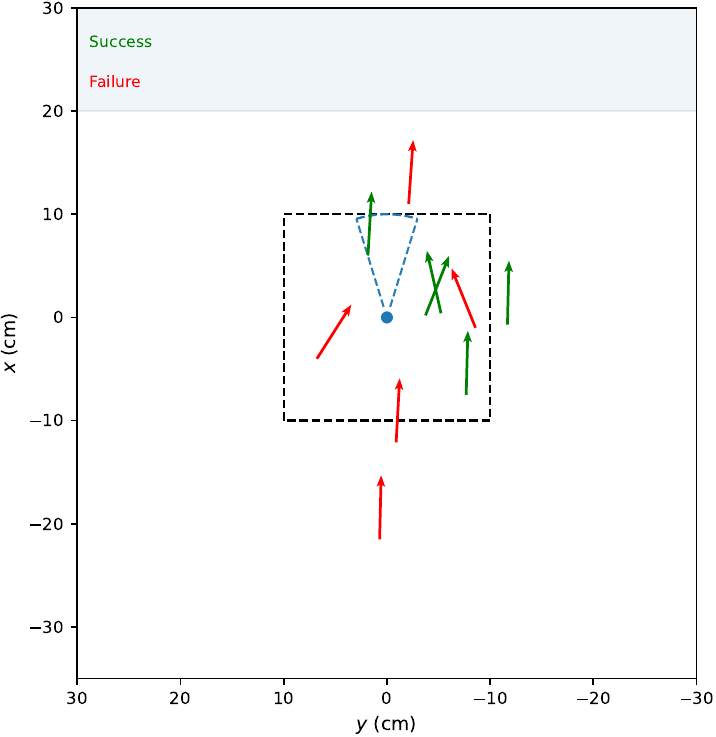}
    \caption{\textbf{Box climb up.}}
  \end{subfigure}%
  \begin{subfigure}[t]{0.49\columnwidth}
    \centering
    \includegraphics[width=\columnwidth]{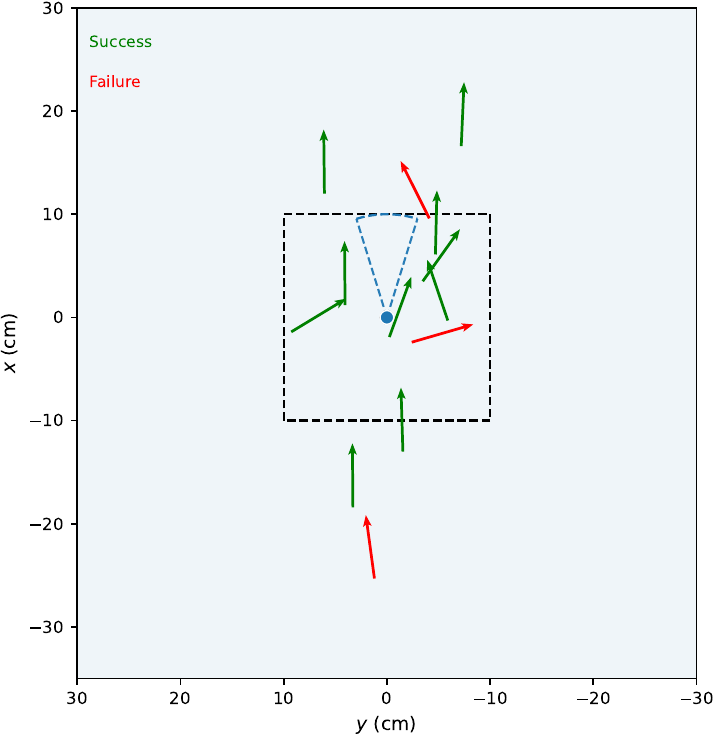}
    \caption{\textbf{Box climb down.}}
  \end{subfigure}

  \caption{
  \textbf{Robustness maps for box-climbing policies from varied initial conditions (top-down view).}
  Each trial is drawn at its initial planar offset $(\Delta x,\Delta y)$ relative to the box, with heading indicated by a short arrow oriented by the initial yaw $\Delta\psi$; green indicates success and red indicates failure.
  Axes follow the manuscript convention: $x$ points upward and $y$ points left.
  The box footprint is shown as a light rectangle (80\,cm along \(x\) and 120\,cm along \(y\)); in this zoomed view, only the portion of the footprint within the plotted window is visible.
  The dashed black square indicates the training randomization range $\Delta x,\Delta y \in [-10,10]$\,cm, and the dashed blue arc indicates yaw randomization $\Delta\psi \in [-0.3,0.3]$\,rad.
  \textbf{Left}: nominal robot start is defined at $(0,0)$ with the box center located 60\,cm ahead along $+x$.
  \textbf{Right}: nominal robot start is aligned with the box center at $(0,0)$. For both box climb up and box climb down, all trials initialized within the training distribution succeed. We also observe successes outside this range (mainly for box climb down), while failures occur only for initial conditions beyond the training distribution. \emph{It is important to note that for visualization purposes, the $y$-positions were scattered by a small offset within $\pm 8$\,cm to reduce overlap (policy is inherently invariant to $y$).}
  }
  \label{fig:box_climb_robustness}
\end{figure}

\newpage
\begin{table}[H]
\centering
\small
\caption{Library of reference motions used for multi-skill policy evaluation and ablation studies.}
\vspace{0.4em}
\label{tab:ref-motions}
\begin{tabular}{c l l S[table-format=2.3, round-mode=places, round-precision=3]}
\toprule
\textbf{Trajectory Index} & \textbf{Motion / Description} & \textbf{Data Source} & \textbf{Duration (s)} \\
\midrule
0  & Army Crawl                     & MoCap     & 16.633333206176758 \\
1  & Dance                          & ViCap     & 9.333240509033203  \\
2  & Stylish Walk                   & ViCap     & 11.533218383789062 \\
3  & Soccer Kick                    & ViCap     & 6.633267402648926  \\
4  & Breakdance                     & MoCap     & 8.300000190734863  \\
5  & Cartwheel                      & MoCap     & 5.9666666984558105 \\
6  & Crouch Walk                    & MoCap     & 8.633333206176758  \\
7  & Crawl on all Fours             & MoCap     & 15.050000190734863 \\
8  & Deep Squat                     & Animation & 2.991666555404663  \\
9  & Animated Walk                  & Animation & 7.483333110809326  \\
10 & Kneeling                       & MoCap     & 8.383333206176758  \\
11 & Run                            & MoCap     & 6.474999904632568  \\
12 & Lightsaber Routine             & Animation & 9.399999618530273  \\
13 & Cartwheel\textendash Backflip  & MoCap     & 5.866666793823242  \\
14 & Roll on all Fours              & MoCap     & 8.800000190734863  \\
\bottomrule
\end{tabular}
\end{table}


\begin{table}[!ht]
    \centering
    \caption{Normalized mean-squared errors of simplified PLA models. Errors are calculated as the differences of the joint accelerations between simplified models and the exact model, and are normalized over the joint accelerations of the exact model.}
    \begin{tabular}{ccc}
        \hline
        & Ankle Pitch  & Ankle Roll\\
        \hline
        \rowcolor[HTML]{EFEFEF} \textit{Locally Projected Model} & 9.527$e^{-4}$ & 3.418$e^{-4}$  \\
        \hline
        \textit{Dynamic Armature Model} & 1.296$e^{-3}$ & 7.902$e^{-4}$  \\
        \hline
        \rowcolor[HTML]{EFEFEF} \textit{Nominal Armature Model} & 1.220$e^{-3}$ & 9.391$e^{-4}$\\
        \hline 
        \textit{Simplest Model} & 0.0776 & 0.8673\\
        \hline
    \end{tabular}    
    \label{tab:actuator_model_approx}
\end{table}

\begin{table}[!ht]
    \centering
    \caption{Observation Terms Summary. We do not perform any scaling or clipping on individual observation terms. All noise models are zero-mean Gaussian and additive in nature. Privileged information is used by the critic only.}
    \vspace{0.5em}
    \label{tab:obs}
    \begin{tabular}{lccc}
        \toprule
        \textbf{Term Name} & \textbf{Definition} & \textbf{Noise} & \textbf{Dim.} \\
        \midrule
        \multicolumn{4}{c}{\hspace{6em}\textbf{Proprioceptive Observations}} \\
        \midrule
        \rowcolor[HTML]{EFEFEF} Torso angular velocity
            & ${}_T \bm \omega_{IT}$
            & $\mathcal{N}(0,\,0.10^2)$
            & $3$ \\
        Gravity vector in torso frame
            & ${}_T \bm g_I$
            & $\mathcal{N}(0,\,0.015^2)$
            & $3$ \\
        \rowcolor[HTML]{EFEFEF} Joint positions
            & $\bm q_j$
            & $\mathcal{N}(0,\,0.005^2)$
            & $n_j$ \\
        Joint velocities
            & $\dot{\bm q}_j$
            & $\mathcal{N}(0,\,0.25^2)$
            & $n_j$ \\
        \rowcolor[HTML]{EFEFEF} Previous action
            & $\bm a_{t-1}$
            & $-$
            & $n_a$ \\
        \midrule
        \multicolumn{4}{c}{\hspace{6em}\textbf{Reference Observations}} \\
        \midrule
        Base height
            & ${}_I \hat{\bm r}^z_{IB}$
            & --
            & $1$ \\
        \rowcolor[HTML]{EFEFEF} Base linear velocity
            & ${}_B \hat{\bm v}_{IB}$
            & --
            & $3$ \\
        Base angular velocity
            & ${}_B \hat{\bm \omega}_{IB}$
            & --
            & $3$ \\
        \rowcolor[HTML]{EFEFEF} Gravity vector in base frame
            & ${}_B \hat{\bm g}_{I}$
            & --
            & $3$ \\
        Joint positions
            & $\hat{\bm q}_j$
            & --
            & $n_j$ \\
        \midrule
        \multicolumn{4}{c}{\hspace{6em}\textbf{Privileged Information (critic only)}} \\
        \midrule
        \rowcolor[HTML]{EFEFEF} Base linear velocity
            & ${}_B \bm v_{IB}$
            & --
            & $3$ \\
        Base height
            & ${}_I \bm r^z_{IB}$
            & --
            & $1$ \\
        \rowcolor[HTML]{EFEFEF} Base contact force
            & ${}_B \bm f_{\text{contact}}$
            & --
            & $3$ \\
        Keybodies contact forces
            & $\{\bm f_{i}\}_{i=1}^{n_{kb}}$
            & --
            & $3 \times n_{kb}$ \\
        \rowcolor[HTML]{EFEFEF} Keybody positions w.r.t. base
            & ${}_B \bm r_{BK}$
            & --
            & $3 \times n_{kb}$ \\
        Keybody linear velocities
            & ${}_B \bm v_{IK}$
            & --
            & $3 \times n_{kb}$ \\
        \rowcolor[HTML]{EFEFEF} Fictitious force (assistive wrench)
            & $\bm f_{\text{assist}}$
            & --
            & $3$ \\
        Fictitious torque (assistive wrench)
            & $\bm \tau_{\text{assist}}$
            & --
            & $3$ \\
        \rowcolor[HTML]{EFEFEF} Wrench scale
            & $\beta$
            & --
            & $1$ \\
        Tracking rewards
            & $\bm r_{\text{track}}$
            & --
            & $7$ \\
        \rowcolor[HTML]{EFEFEF} Task phase
            & $\phi$
            & --
            & $1$ \\
        \bottomrule
    \end{tabular}
\end{table}

\begin{table}[t]
    \centering
    \caption{Reward Terms Summary. The environment scales the reward weights with the time-step $dt$. For brevity, the time index $t$ is omitted unless needed. Keybody positions and orientations are expressed w.r.t.\ the base frame. A common stiffness parameter of $\kappa = 1/4$ is set for all tracking reward terms.}
    \label{tab:rewards}
    \vspace{0.6em}
    \begin{tabular}{lccc}
        \toprule
        \textbf{Term Name} & \textbf{Definition} & \textbf{Weight} & \textbf{$\sigma_i$} \\
        \midrule
        \multicolumn{4}{c}{\hspace{6em}\textbf{Tracking Reward}} \\
        \midrule
        \rowcolor[HTML]{EFEFEF} Base position tracking
            & $\exp\!\bigl(-\kappa\,\|\mathbf{r}_{IB} - \mathbf{r}^*_{IB}\|^2 / \sigma_1^2\bigr)$
            & $1$ & $0.4$ \\
        Base orientation 
            & $\exp\!\bigl(-\kappa\,\|\bm{\Phi}_{IB} \boxminus \bm{\Phi}^*_{IB}\|^2 / \sigma_2^2\bigr)$
            & $1$ & $0.5$ \\            
        \rowcolor[HTML]{EFEFEF} Base angular velocity 
            & $\exp\!\bigl(-\kappa\,\|\bm{\omega}_{IB} - \bm{\omega}^*_{IB}\|^2 / \sigma_3^2\bigr)$
            & $1$ & $1.5$ \\
        Base linear velocity 
            & $\exp\!\bigl(-\kappa\,\|\mathbf{v}_{IB} - \mathbf{v}^*_{IB}\|^2 / \sigma_4^2\bigr)$
            & $1$ & $0.6$ \\
        \rowcolor[HTML]{EFEFEF} Joint position 
            & $\exp\!\bigl(-\kappa\,\|\bm{q}_j - \bm{q}^*_j\|^2 / \sigma_5^2\bigr)$
            & $1$ & $0.3\cdot\sqrt{n_j}$ \\
        \rowcolor[HTML]{EFEFEF} Keybodies position 
            & $\exp\!\bigl(-\kappa\,\|\bm{r}_{kb} - \bm{r}^*_{kb}\|^2 / \sigma_6^2\bigr)$
            & $1$ & $0.2\cdot\sqrt{n_{kb}}$ \\
        Keybodies orientation 
            & $\exp\!\bigl(-\kappa\,\|\bm{\Phi}_{kb} \boxminus \bm{\Phi}^*_{kb}\|^2 / \sigma_7^2\bigr)$
            & $1$ & $0.4\cdot\sqrt{n_{kb}}$ \\
        \midrule
        \multicolumn{4}{c}{\hspace{6em}\textbf{Regularization Penalty}} \\
        \midrule
        Action smoothness
            & $-\|\bm{a}_t - \bm{a}_{t-1}\|$
            & $0.15$ & -- \\
        \rowcolor[HTML]{EFEFEF} Joint acceleration
            & $-\|\ddot{\bm q}_j\|$
            & $1\times 10^{-5}$ & -- \\
        Joint position limit
            & $-\displaystyle \sum_{i \in \mathcal{J}}\!\Bigl[\max(0, q^{\min}_i - q_i) + \max(0, q_i - q^{\max}_i)\Bigr]$
            & $1.0$ & -- \\
        \rowcolor[HTML]{EFEFEF} Joint torque limit
            & $-\displaystyle \sum_{i \in \mathcal{J}}\!\Bigl[\max(0, \tau^{\min}_i - \tau_i) + \max(0, \tau_i - \tau^{\max}_i)\Bigr]$
            & $0.1$ & -- \\
        \midrule
        \multicolumn{4}{c}{\hspace{6em}\textbf{Survival Reward}} \\
        \midrule
        \rowcolor[HTML]{EFEFEF} Survival
            & $1$
            & $1$ & -- \\
        \bottomrule
    \end{tabular}
\end{table}

\begin{table}[ht]
\centering
\caption{Domain randomization terms including dynamics randomization and external perturbations.}
\label{tab:domain-rand}
\vspace{0.6em}
\begin{tabular}{l|l}
\toprule
\textbf{Term} & \textbf{Value} \\
\midrule
Static friction      & $\mathcal{U}(0.6,\;1.0)$ \\
\rowcolor[HTML]{EFEFEF} Dynamic friction      & $\mathcal{U}(0.5,\;0.9)$ \\
Restitution      & $\mathcal{U}(0.0,\;0.2)$ \\
\rowcolor[HTML]{EFEFEF} Link masses     & $\mathcal{U}(0.9,\;1.1)\,\times$ default masses \\
External disturbance (impulsive push applied to base) & Interval $=\mathcal{U}(\SI{0}{\second},\;\SI{10}{\second})$,\quad $v_{xy}=\SI{0.5}{\meter\per\second}$ \\
\bottomrule
\end{tabular}
\end{table}

\begin{table}[ht]
\centering
    \caption{MDP Hyperparameters.}
    \label{tab:adp}
    \begin{tabular}{l|c} 
        \toprule
        \textbf{Hyperparameter} & \textbf{Value} \\
        \midrule
            Episode length $(\bar{L}_{\mathrm{episode}})$ & \SI{10}{\second} \\
            \rowcolor[HTML]{EFEFEF} Simulation time-step $(dt)$ & \SI{0.004}{\second} \\
            Control decimation & 5 \\
            \rowcolor[HTML]{EFEFEF} Action scale (per joint) & $\{0.05,\;0.10,\;0.20\}$ \\
            Bin width $(\Delta)$ & $\min\!\bigl(\SI{4.0}{\second},\, \min_i D_i\bigr)$ \\
            \rowcolor[HTML]{EFEFEF} EMA alpha $(\alpha)$ & 0.005 \\
            Base temperature $(\tau_{\text{base}})$ & 1.0 \\
            \rowcolor[HTML]{EFEFEF} Adaptive sampling temperature $(\tau)$ & $\displaystyle \frac{\tau_{\text{base}}}{\log\!\bigl(1+\lvert\Omega\rvert\bigr)}$ \\
            Uniform floor weight $(\varepsilon)$ & 0.15 \\ 
            \rowcolor[HTML]{EFEFEF} Max wrench scale $(\beta_{\max})$ & 0.60 \\
            Similarity threshold $(\eta)$ & 0.80 \\
            \rowcolor[HTML]{EFEFEF} Virtual torque PD gains $(k_p^{\omega},\, k_d^{\omega})$ & $(200.0,\, 10.0)$ \\
            Virtual force PD gains $(k_p^{v},\, k_d^{v})$ & $(0.0,\, 10.0)$ \\
        \bottomrule
    \end{tabular}
\end{table}

\begin{table}[ht]
\centering
    \vspace{-13pt}
    \caption{PPO Hyperparameters}
    \label{tab:ppo-hp}
    \begin{tabular}{l|c} 
        \toprule
        \textbf{Hyperparameter} & \textbf{Value} \\
        \midrule
                                Actor Network & MLP(512, 256, 128) with \emph{ELU} activation \\
        \rowcolor[HTML]{EFEFEF} Critic Network & MLP(512, 512, 256) with \emph{ELU} activation \\
                                Empirical Normalization & True \\
        \rowcolor[HTML]{EFEFEF} Learning Rate (start of training) & 1e-3 \\
                                Learning Rate Schedule & ``adaptive"  (based on KL-divergence)\\
        \rowcolor[HTML]{EFEFEF} Discount Factor & 0.99 \\
                                GAE Discount Factor        & 0.95 \\
        \rowcolor[HTML]{EFEFEF} Desired KL-divergence & 0.01 \\
                                Clip Range & 0.2 \\
        \rowcolor[HTML]{EFEFEF} Entropy Coefficient & 0.001 \\
                                Value Function Loss Coefficient & 0.5 \\
        \rowcolor[HTML]{EFEFEF} Number of Epochs & 5 \\
                                Number of Environments & 4096 \\
        \rowcolor[HTML]{EFEFEF} Batch Size & 245,760 $(4096 \times 24)$ \\
                                Mini-Batch Size & 61,440 $(4096 \times 6)$ \\       
        \bottomrule
    \end{tabular}
\end{table}

\end{document}